%% file: main.tex
\documentclass{article}

\usepackage{PRIMEarxiv}

\usepackage[utf8]{inputenc}
\usepackage[T1]{fontenc}
\usepackage{hyperref}
\usepackage{url}
\usepackage{booktabs}
\usepackage{amsmath}
\usepackage{amsfonts}
\usepackage{nicefrac}
\usepackage{microtype}
\usepackage{graphicx}
\usepackage{subcaption}
\usepackage[table]{xcolor}
\usepackage{listings}
\usepackage{float}
\usepackage[section]{placeins}
\usepackage{tikz}
\usepackage{enumitem}
\graphicspath{{./}{assets/}{assets/logo/}}
\renewcommand{\headername}{AVTR-1}
\renewcommand{\headerright}{Technical Report}

\definecolor{codebg}{RGB}{248,248,248}
\definecolor{codekw}{RGB}{0,92,175}
\definecolor{codecom}{RGB}{0,128,0}
\definecolor{codestr}{RGB}{163,21,21}
\definecolor{bestgray}{gray}{0.72}
\definecolor{secondgray}{gray}{0.90}
\definecolor{significantgray}{gray}{0.82}

\lstdefinestyle{cleanpseudo}{
  backgroundcolor=\color{codebg},
  basicstyle=\ttfamily\small,
  keywordstyle=\color{codekw}\bfseries,
  commentstyle=\color{codecom},
  stringstyle=\color{codestr},
  numbers=left,
  numberstyle=\tiny,
  numbersep=8pt,
  frame=single,
  breaklines=true,
  showstringspaces=false,
  tabsize=2
}

\newcommand{\roundedtitlelogo}[1]{%
  \tikz[baseline=(logo.base)] \node[rounded corners=3pt, inner sep=0pt, clip] (logo)
  {\includegraphics[width=1.1cm,height=1.1cm]{#1}};%
}
\newcommand{\roundedinlinelogo}[2][1.1em]{%
  \tikz[baseline=(logo.base)] \node[rounded corners=1.5pt, inner sep=0pt, clip] (logo)
  {\includegraphics[height=#1]{#2}};%
}

\title{\resizebox{\textwidth}{!}{\raisebox{-0.30\height}{\roundedtitlelogo{\detokenize{avaturn_logo.png}}}\hspace{0.55em}AVTR-1: Open Stack for Real-Time Interactive Avatars}}

\author{\normalfont
  Artem Kravtsov\textsuperscript{*}, Dmitrii Ziganshin\textsuperscript{*}, Vsevolod Poletaev\textsuperscript{*},\\
  Gleb Balitskiy, Anastasia Tikhonova, Egor Burkov, Vadim Lebedev\\[0.8em]
  Avaturn Live\\
  \textsuperscript{*}Core contributors: co-wrote the paper
}
\date{}

\begin{document}
\maketitle

\begin{abstract}
  Talking-head and dyadic models now achieve real-time inference, yet fast motion generation alone does not produce an
  interactive conversation. A live system must synchronize the model's output with speech from an external voice agent,
  schedule video frames for playback, and handle interruptions. We introduce AVTR-1, an open stack for real-time
  interactive avatar conversations, built around a compact 153M-parameter autoregressive flow-matching motion generator
  conditioned on both participants' audio. We adapt its audio encoder for streaming through self-distillation. The stack
  turns the model's chunk-based generation into a continuous, synchronized audio-video stream driven by an external
  voice agent, and we analytically derive its contribution to the user-facing latencies and validate the resulting
  bounds with two commercial voice agents. Further experiments demonstrate that AVTR-1 leads the compared dyadic systems
  on all reported visual-quality metrics and most conventional listening-motion metrics while remaining competitive in
  lip synchronization. Its inference runtime operates in real time on data-center and consumer GPUs. However,
  conventional listening metrics do not establish whether the paired speaker's speech contributes to generated motion.
  We therefore introduce the Reference-Based Directed Granger Gain (R-DGG), which measures the additional predictive
  information carried by speaker speech after accounting for listener history and speaker motion. R-DGG finds
  statistically supported predictive dependence for recorded listeners and all evaluated dyadic systems, but not for
  talking-head generators without paired audio or mismatched speaker--listener pairs. We release the model weights,
  renderer, and serving backend under component-specific licenses. \\[0.8em]
  \raisebox{-0.2em}{\roundedinlinelogo{\detokenize{avaturn_logo.png}}}~\href{https://avaturn.live}{https://avaturn.live}\\[0.3em]
  \raisebox{-0.2em}{\roundedinlinelogo{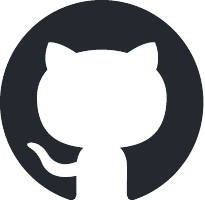}}~\href{https://github.com/avaturn-live/avtr-1}{https://github.com/avaturn-live/avtr-1}\\[0.3em]
  \raisebox{-0.2em}{\includegraphics[height=1.1em]{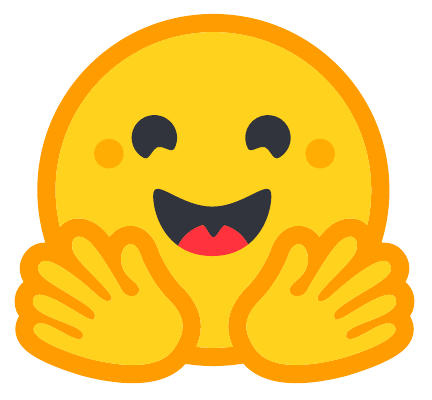}}~\href{https://huggingface.co/avaturn-live/avtr-1}{https://huggingface.co/avaturn-live/avtr-1}
\end{abstract}

\newpage
\tableofcontents
\newpage

\input{sections/introduction}

\section{Motion Generation Model}
\label{sec:motion-generation-model}
\input{sections/motion_representation}
\input{sections/architecture}

\section{Model Training}
\label{sec:model-training}
\input{sections/data}
\input{sections/training}

\section{Interactive Real-Time Streaming Stack}
\label{sec:real-time-inference-serving}
\input{sections/hubert}
\input{sections/renderer}
\input{sections/streamer}
\input{sections/latency}
\input{sections/evaluation}
\input{sections/conclusion}

\section*{Acknowledgments}
We thank Sergei Sherman for supporting this work and providing feedback.

\FloatBarrier

\bibliographystyle{unsrt}
\bibliography{references}

\appendix
\input{sections/appendix}

\end{document}

%% file: sections/introduction.tex
\section{Introduction}
\label{sec:intro}

Speech-driven talking-head generation has advanced rapidly, producing increasingly natural and identity-faithful video
from speech and a reference portrait~\cite{li2024ditto,ki2024float,soulx}. Recent work has begun moving from offline
generation toward real-time dyadic systems~\cite{guo2025arig,ki2026avatarforcing,chen2025dystream}. However, fast model
inference alone is insufficient for live conversation, which requires both speech-conditioned listening and timely
delivery of synchronized audio and video.
Speech from an external voice agent arrives in variable-sized fragments, whereas a chunk-based motion generator requires
fixed-size audio windows. The surrounding live system must adapt one to the other while continuously generating new
frames, synchronizing them with audio, scheduling them for playback, and handling interruptions. Its scheduling
decisions shape two user-facing latencies: the response latency, how soon the avatar's reply becomes audible, and the
interruption latency, how long speech continues after the user interrupts.

We introduce AVTR-1, an open stack for interactive avatar conversations spanning motion generation, inference, and
real-time serving. The stack connects to an external voice agent for the conversation itself: it leaves the
interpretation of user speech and the generation of the avatar's reply speech to that agent, and consumes the agent's
output. We call the inference-side component the renderer and the serving backend the streamer. The motion generator
predicts head and facial motion from both participants' audio. The renderer converts predicted motion into video and
returns updated session state. The streamer coordinates the live session and drives the renderer in a continuous loop.
Together, these components let us examine generation and serving within one system.

AVTR-1 separates motion synthesis from appearance rendering to keep the generation target compact. A 153M-parameter
conditional flow-matching Transformer predicts five-frame motion chunks autoregressively, using self audio for speaking
motion and other audio for listening behavior. We train the model on an internal corpus of 926 hours of curated,
unscripted dyadic conversations, gradually replacing ground-truth motion history with model estimates to match
autoregressive inference. Training uses audio features precomputed from full-length recordings, but live inference has
access only to short chunks of audio. We address this mismatch by self-distilling the audio encoder to reproduce its
full-context features from short windows.

The streamer converts variable-sized speech fragments from the voice agent to the fixed-size audio windows consumed by
the renderer. To support an interactive video call, frames must be generated and played continuously, even when the
voice agent produces no speech. The streamer drives this continuous generation, forms the audio windows required by each
request, and delivers the resulting synchronized audio-video stream to the user. We derive the streamer's contribution
to both latencies from its scheduling rules and validate the resulting bounds with two commercial voice agents.

We evaluate visual quality, speaking quality, and reactive listening against two dyadic models and four talking-head
generators. AVTR-1 achieves leading performance among the compared dyadic models across visual quality, lip
synchronization, and listening behavior, while its inference runtime operates in real time on data-center and consumer
GPUs. Listening evaluation presents a separate difficulty: a generated listener can produce plausible motion without
responding to the paired speaker, and motion-based metrics do not establish this dependence. We therefore introduce the
Reference-Based Directed Granger Gain (R-DGG), which measures the additional predictive information from speaker speech
after accounting for listener history and speaker motion. Our experiments demonstrate that R-DGG indicates speech
dependence for ground-truth listener motion and dyadic-model outputs, but not for two types of negative controls:
talking-head generators and mismatched speaker--listener pairs.

We release the model weights, renderer, and streamer under component-specific licenses. Together, these artifacts cover
the complete path required to run a live interactive avatar session locally. The remainder of this report describes the
motion model (Section~\ref{sec:motion-generation-model}) and training pipeline (Section~\ref{sec:model-training}), the
streaming inference and serving stack (Section~\ref{sec:real-time-inference-serving}), the latency analysis
(Section~\ref{sec:latency}), and the evaluation protocol and results (Section~\ref{sec:evaluation}).

%% file: sections/motion_representation.tex
\subsection{Motion Representation}
\label{sec:motion}

For each frame, AVTR-1 generates a 42-dimensional vector of head rotations and expression deformations. This
motion representation is derived from LivePortrait's motion--appearance disentangled space~\cite{guo2024liveportrait}.
Unlike a VAE latent that entangles the two, this compact representation
reduces training and sampling cost while enabling per-region control.

\paragraph{Motion parameters.} LivePortrait implements this decomposition with motion and appearance extractors.
The motion extractor predicts canonical keypoints ($\mathbf{x}_c \in \mathbb{R}^{21 \times 3}$), head pose
($\mathbf{R} \in \mathrm{SO}(3)$), expression deformations
($\boldsymbol{\delta} \in \mathbb{R}^{21 \times 3}$), scale ($s \in \mathbb{R}_{>0}$), and translation
($\mathbf{t} \in \mathbb{R}^{3}$). The transformed keypoints are computed as follows:
\begin{equation}
  \mathbf{x}_d = s(\mathbf{x}_c\mathbf{R} + \boldsymbol{\delta}) + \mathbf{t}.
  \label{eq:lp-transform}
\end{equation}
The appearance extractor produces the volumetric feature used for rendering. We construct the training targets
with the motion extractor alone, since appearance is not part of the training target. Further details are
provided in the original work.

\paragraph{Training target.} AVTR-1 uses a reduced subset of these parameters for its per-frame target.
LivePortrait represents expression as 3D deformations of 21 keypoints, yielding 63
coordinates. We choose the 39 coordinates that correspond to the brow, eyes, and mouth regions. Before
selecting them, we transform deformations into the head coordinate frame as
$\tilde{\boldsymbol{\delta}} = \boldsymbol{\delta}\mathbf{R}^{\mathsf{T}}$. For the rotation, we take the
axis-angle vector $\boldsymbol{r} \in \mathbb{R}^{3}$ of $\mathbf{R}$. Let $z_r$ and $z_\delta$ denote
per-coordinate z-score normalization using dataset-global statistics. The training target is then
\begin{equation}
  \mathbf{m} \;=\; \operatorname{concat}\big[\, z_r(\boldsymbol{r}) \;;\;
  z_{\delta}(\operatorname{vec}(\tilde{\boldsymbol{\delta}})_{\mathcal{I}}) \,\big] \;\in\; \mathbb{R}^{42},
  \label{eq:motion-target}
\end{equation}
where $\operatorname{vec}$ flattens the $21\times 3$ deformation to a 63-vector and $\mathcal{I}$ selects the
fixed set of 39 selected expression coordinates.

We exclude the scale, the translation, the canonical keypoints, and the 24 remaining expression coordinates
from the training target. At render time they are taken from the source image and held fixed.

\paragraph{Reference condition.} The motion generation Transformer also receives a static reference condition. During training, we construct it
for each person track by concatenating the medians of the rotation vector, all 63 canonical keypoint
coordinates, and all 63 expression coordinates in the head coordinate frame. During inference, it is
constructed from the source image. Section~\ref{sec:architecture} describes how the model encodes it.

%% file: sections/architecture.tex
\subsection{Model Architecture}
\label{sec:architecture}

\begin{figure}[t]
  \centering
  \includegraphics{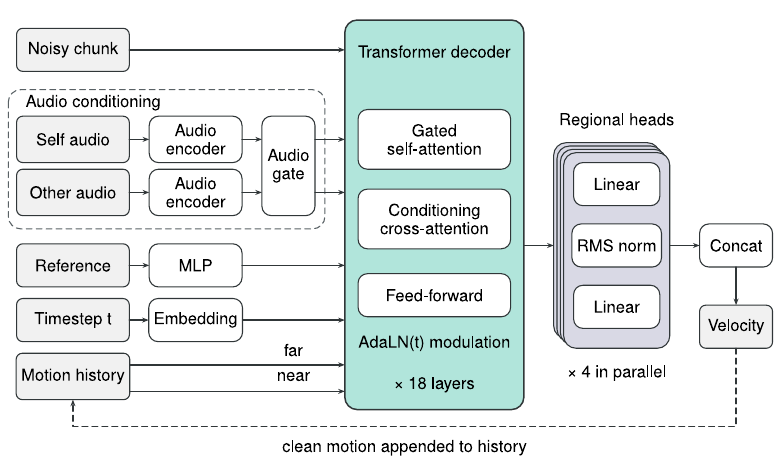}
  \caption{Overview of the AVTR-1 motion model.}
  \label{fig:architecture}
\end{figure}

The AVTR-1 motion model is a 153M-parameter conditional flow-matching Transformer~\cite{lipman2023flow} that
autoregressively generates five-frame motion conditioned on two separate audio channels called self and other audio.
Figure~\ref{fig:architecture} shows the model architecture, and Table~\ref{tab:arch} lists its configuration.

\begin{table}[t]
  \centering
  \caption{Configuration of the released AVTR-1 motion model.}
  \label{tab:arch}
  \begin{tabular}{ll}
    \toprule
    Property & Value \\
    \midrule
    Parameters & 153M \\
    Decoder layers & 18 \\
    Hidden dimension & 512 \\
    Attention heads / head dimension & 8 / 64 \\
    Feed-forward dimension & 512 \\
    Audio condition encoder layers & 2 per stream \\
    Dropout & 0 \\
    RoPE base & $10^4$ \\
    \midrule
    Generated motion chunk & 5 frames \\
    Past motion context & 75 frames \\
    Motion dimension & 42 \\
    \midrule
    Audio context per stream (past / present / future) & 75 / 5 / 5 feature steps \\
    Audio feature dimension per stream & 1024 \\
    Reference condition & 129 \\
    \bottomrule
  \end{tabular}
\end{table}

\paragraph{Audio encoder.}
We use the 24-layer pre-trained HuBERT Large encoder, with roughly 300M parameters. The audio encoder produces
1024-dimensional features at 50\,Hz. We average adjacent pairs to match the motion model's 25\,fps rate. We extract
features for two audio channels separately.

\paragraph{Transformer.} The motion model is a pre-norm Transformer decoder with 18 layers, 8 heads, and width 512. Each
layer applies three residual sub-blocks: self-attention, a conditioning block, and a feed-forward block. We restrict
self-attention to the five-frame chunk and route the past motion and audio through the conditioning block. This keeps
the cost per chunk independent of how much motion has already been generated. The flow-matching timestep modulates each
sub-block through adaptive layer normalization (AdaLN), with separate scale and shift predicted for each. We
zero-initialize the corresponding projections, so AdaLN initially reduces to standard LayerNorm. The noisy current chunk
and the past motion history share a linear projection from 42 to 512 dimensions. Following work on attention
stability~\cite{dehghani2023vit22b}, we apply RMSNorm~\cite{zhang2019rmsnorm} to queries and keys before computing
attention scores. Unlike QK normalization that shares scale parameters across heads~\cite{esser2024sd3}, our formulation
learns separate query and key scale vectors for every head. Each scale vector is parameterized as $(1+\mathbf{w})$, with
$\mathbf{w}$ initialized to zero. Every head therefore starts at unit gain, while weight decay regularizes its gain
toward one rather than zero. The normalization controls attention logit growth while allowing each head to adjust its
scale independently. Additionally, each self-attention applies a head-specific learned sigmoid gate to its output,
initialized with zero weights and a bias of 2. This design follows the gated-attention scheme of Qiu et
al.~\cite{qiu2025gated}, which was introduced to mitigate attention sinks. All attention modules use the same
RoPE~\cite{su2024roformer} configuration, while positions are constructed independently for each query and key sequence.

\paragraph{Conditioning.} We route the conditions through separate paths rather than fusing them in a single attention
operation. Each block receives past motion, audio features for both channels, and a static reference. We divide the
75-frame motion history between a near path over the last five frames and a far path over the earlier 70. All paths use
cross-attention except the reference, which uses an MLP. This separation prevents audio tokens from competing with
motion and reference inputs in the same attention distribution. Let $\mathbf{A}_{\mathrm{far}}^{(\ell)}$,
$\mathbf{A}_{\mathrm{near}}^{(\ell)}$, $\mathbf{A}_{\mathrm{self}}^{(\ell)}$, and $\mathbf{A}_{\mathrm{other}}^{(\ell)}$
denote the four cross-attention outputs in decoder layer $\ell$, and let $\mathbf{R}^{(\ell)}$ denote the reference MLP
output added to each token of the current chunk. The binary condition-availability vector
$\mathbf{b}=(b_{\mathrm{past}},b_{\mathrm{self}},b_{\mathrm{other}},b_{\mathrm{ref}})\in\{0,1\}^{4}$ indicates which
conditions are retained. The masked outputs form the conditioning residual $\Delta\mathbf{H}_{\mathrm{cond}}^{(\ell)}$,
which is added to the hidden state after self-attention:
\begin{equation}
  \begin{aligned}
    \Delta\mathbf{H}_{\mathrm{cond}}^{(\ell)}={}&
    b_{\mathrm{past}}\left(\mathbf{A}_{\mathrm{far}}^{(\ell)}+\mathbf{A}_{\mathrm{near}}^{(\ell)}\right) \\
    &+b_{\mathrm{self}}\mathbf{A}_{\mathrm{self}}^{(\ell)}
    +b_{\mathrm{other}}\mathbf{A}_{\mathrm{other}}^{(\ell)}
    +b_{\mathrm{ref}}\mathbf{R}^{(\ell)}.
  \end{aligned}
  \label{eq:cond-residual}
\end{equation}
The near and far history paths share $b_{\mathrm{past}}$, so guidance treats past motion as a single condition.

Since automatic speaker separation in the training data is imperfect, we introduce an audio gate with direct
speaker-activity supervision. Following the complementary weighting used in the Gated Multimodal
Unit~\cite{arevalo2017gmu}, the gate predicts channel-wise weights jointly from both streams but keeps the weighted
representations separate. During training, we average the gate logits across channels and supervise the result with
pseudo-VAD labels on frames where a single speaker is confidently active.

\paragraph{Regional output heads.} The Transformer decoder produces five 512-dimensional tokens. We apply four linear
projections to each token to create separate latent spaces for head rotation, brow, eyes, and mouth. Each projection
operates on the full 512-dimensional representation, while its output dimension is proportional to the number of motion
coordinates in that region. This intermediate split allows classifier-free guidance (CFG) to be applied independently to
the four regions. Each regional latent is then normalized by its RMS across the five frames and latent channels. Four
additional linear layers map each regional latent to the corresponding motion coordinates. The outputs are concatenated
to form the final 42-dimensional velocity prediction.

%% file: sections/data.tex
\subsection{Data Pipeline}
\label{sec:data}

AVTR-1 is trained on an internal corpus of dyadic conversational videos that is not included in this release.
We construct it by progressively segmenting, filtering, tracking, separating, and annotating a larger collection
of candidate videos, retaining 926 hours of unscripted conversation. Each retained segment contains synchronized
left and right participant tracks, separate audio tracks, and LivePortrait motion annotations. Every pipeline
stage saves its intermediate outputs. This lets
us schedule CPU and GPU workloads separately, identify bottlenecks, and revise individual components without
repeating earlier steps.

\begin{figure}[t]
  \centering
  \includegraphics{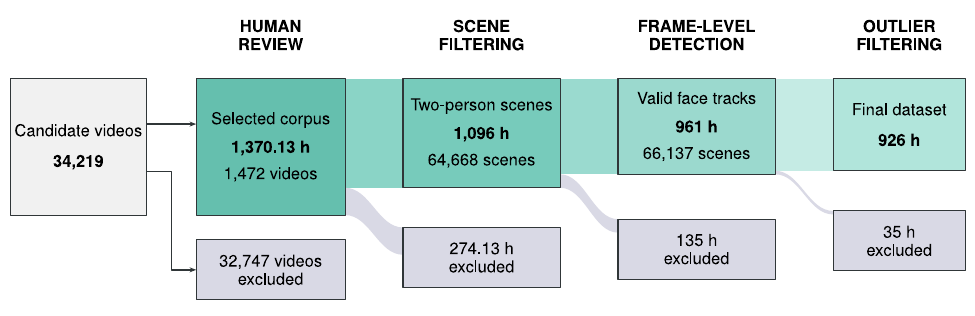}
  \caption{Data filtering cascade.}
  \label{fig:data-filtering}
\end{figure}

Figure~\ref{fig:data-filtering} summarizes the data retained after each filtering stage.

\paragraph{Source video selection.} The source collection contained 34{,}219 candidate videos. Human annotators
retained 1{,}472 videos, totaling 1{,}370.13 hours. Each retained video contained at least one segment in
which exactly two visible participants conversed with each other, with one speaking and the other listening.
Both faces had to be clearly visible from the front or at a slight angle and remain free of occlusion. Annotators
excluded short-form videos, edited excerpts of already selected videos, and uncertain cases.

\paragraph{Scene detection.} We split each retained video into scenes with PySceneDetect~\cite{pyscenedetect},
discard the first 90 seconds, and exclude any scene shorter than five seconds from face filtering. Within the
remaining scenes, we sample every 25th frame and run a face detector, keeping faces above a minimum size of
160 pixels at 1080p resolution. When the face detector does not return exactly two faces, for example because it
detects faces in background images, we use a YOLO person detector~\cite{ultralytics_yolo} as a fallback. We retain
segments of at least five seconds in which every
sampled frame passes this two-person check.

\paragraph{Face detection.} For every frame in each segment, we define fixed participant regions over its
left and right halves. A whole-body pose estimator processes both regions, and each participant's 68 facial
landmarks must lie within the corresponding half and have a mean confidence above 0.3. We retain sequences of
frames that satisfy these checks and last at least five seconds. This stage produces frame-level landmark
annotations for the left and right participants, which the later stages consume.

\paragraph{Speaker separation.} The source audio contains both voices on the same track. The motion model
requires a separate stream for each participant. We convert the audio to 16 kHz mono and separate the two
voices with ClearerVoice-Studio's pretrained audiovisual target-speaker extractor~\cite{pan2023avtfgridnet,pan2025isam}.
The model also receives cropped face videos for the left and right participants and uses their lip motion to
associate each output with the corresponding side. During training, we randomly select one participant as the
generation target and treat their waveform as self audio and the other waveform as other audio.

\paragraph{Audio features.} We encode the two separated waveforms independently with the original HuBERT encoder.
Concatenating the two 1024-dimensional streams produces a 2048-dimensional vector
per frame. We store one feature array per segment.

\paragraph{Motion extraction.} The facial landmarks from the preceding stage define a crop for each
participant in every retained frame. We smooth the crop boxes over 25 frames, resize each face to
$256\times256$, and process the left and right crops in the same batch with the LivePortrait motion
extractor~\cite{guo2024liveportrait}. The full extractor output is stored for both participant tracks.
Section~\ref{sec:motion} describes how these annotations are converted into the training target and reference
condition.

\paragraph{Training-window filtering.} We exclude training windows with implausible motion. For
each participant track, we compute first-, second-, and third-order temporal finite differences of the rotation and
expression coordinates. We scale these differences per coordinate, average their absolute values, and smooth
the resulting score over 21 frames. A frame is marked as an outlier when its score exceeds 0.15, and the mask
extends five frames in either direction. We retain only 25-frame windows that do not overlap this mask. A
segment is discarded if either participant track has no windows left.

%% file: sections/training.tex
\subsection{Training Procedure}
\label{sec:training}

We train the AVTR-1 motion model using the conditional flow-matching framework~\cite{lipman2023flow}. Training follows
the autoregressive inference setup: each five-frame chunk is conditioned on earlier motion predictions and two audio
streams. The objective combines a flow-matching velocity loss, temporal smoothness regularizers, and direct supervision
of the audio gate. Training further uses correlated and immiscible noise and multi-condition dropout for guidance.

\paragraph{Audio feature extraction.}
We run the pre-trained HuBERT audio feature extractor once before training on full audio durations and store the results.
This approach lets us sample training windows of different lengths without rerunning the encoder, which increases
training efficiency.

\paragraph{Loss functions.} Given a motion target $\mathbf{m}$ and Gaussian noise $\boldsymbol{\epsilon}$, we construct
a noisy sample at timestep $t\in[0,1]$ by linear interpolation:
\begin{equation}
  \mathbf{m}_t = (1-t)\,\boldsymbol{\epsilon} + t\,\mathbf{m}.
\end{equation}
The model predicts a velocity $\mathbf{v}_\theta(\mathbf{m}_t,t)$. The ground-truth target is the path velocity
$\tfrac{d\mathbf{m}_t}{dt}=\mathbf{m}-\boldsymbol{\epsilon}$. We compute the mean-squared error separately for head
rotation, brow, eyes, and mouth, then sum the four regional losses:
\begin{equation}
  \mathcal{L}_{\mathrm{CFM}}=\mathbb{E}_{t,\mathbf{m},\boldsymbol{\epsilon}}\sum_{r=1}^{4}
  \operatorname{MSE}\left(\mathbf{v}_\theta^{(r)}(\mathbf{m}_t,t),\,\mathbf{m}^{(r)}-\boldsymbol{\epsilon}^{(r)}\right).
\end{equation}
Averaging within each region gives all four regions equal weight regardless of their coordinate count. We sample one
timestep per five-frame chunk from a logit-normal distribution~\cite{esser2024sd3} and share it across the frames of the
chunk, matching the chunkwise generation performed at inference. We also compute the cosine distance
$\mathcal{L}_{\mathrm{cos}}$ between the predicted and target velocities separately for the same four regions. This term
constrains the direction of the predicted flow independently of its magnitude.

To encourage smooth motion, we regularize the expression coordinates, head rotation, and reconstructed 3D keypoint
motion separately. Let $\hat{\mathbf{e}}$ denote the predicted expression coordinates, $\hat{\mathbf{R}}$ the predicted
rotations, and $\hat{\mathbf{q}}$ the selected 3D keypoints obtained by adding $\hat{\mathbf{e}}$ to the reference
canonical keypoints and applying $\hat{\mathbf{R}}$. We define $\boldsymbol{\omega}_t$ as the axis-angle vector of the
relative rotation $\hat{\mathbf{R}}_{t-1}^{\mathsf{T}}\hat{\mathbf{R}}_t$ and minimize
\begin{equation}
\begin{aligned}
\mathcal{L}_{\mathrm{reg}}={}&\lambda_{\mathrm{exp}}\left(\lVert \Delta\hat{\mathbf{e}} \rVert_2^2
+ \lVert \Delta^2\hat{\mathbf{e}} \rVert_2^2 + \lVert \Delta^3\hat{\mathbf{e}} \rVert_2^2\right) \\
&+\lambda_{\mathrm{rot}}\left(\lVert \boldsymbol{\omega} \rVert_2^2 + \lVert \Delta\boldsymbol{\omega} \rVert_2^2
+ \lVert \Delta^2\boldsymbol{\omega} \rVert_2^2\right) \\
&+\lambda_{\mathrm{3D}}\left(\lVert \Delta\hat{\mathbf{q}} \rVert_2^2 + \lVert \Delta^2\hat{\mathbf{q}} \rVert_2^2
+ \lVert \Delta^3\hat{\mathbf{q}} \rVert_2^2\right) ,
\end{aligned}
\end{equation}
where $\Delta$ is the temporal finite-difference operator. The coefficients $\lambda_{\mathrm{exp}}$,
$\lambda_{\mathrm{rot}}$, and $\lambda_{\mathrm{3D}}$ weight the three regularization terms. Each squared norm is
averaged over frames and coordinates. We prepend the last three frames of the motion history before computing these
terms, so the regularization extends across chunk boundaries.

We supervise the audio gate described in Section~\ref{sec:architecture} on high-confidence single-speaker frames
identified by a frozen voice-activity detector. A binary cross-entropy loss $\mathcal{L}_{\mathrm{VAD}}$ on the
channel-averaged gate logit uses a target of 1 when self audio is active and 0 when other audio is active. This
auxiliary loss helps suppress cross-speaker leakage caused by imperfect speaker separation. The total training objective
is
\begin{equation}
  \mathcal{L}_{\mathrm{total}}=\mathcal{L}_{\mathrm{CFM}}+\mathcal{L}_{\mathrm{cos}}+\mathcal{L}_{\mathrm{reg}}
  +\lambda_{\mathrm{VAD}}\mathcal{L}_{\mathrm{VAD}},
\end{equation}
where $\lambda_{\mathrm{VAD}}$ controls the weight of the audio-gate supervision.

\paragraph{Noise design.} We use progressive noise~\cite{ge2023pyoco}, which forms an autoregressive noise trajectory
over the motion sequence. Each frame's noise is correlated with the preceding frame, and the trajectory continues across
five-frame chunk boundaries. During training, we also apply immiscible noise assignment~\cite{li2024immiscible}
independently to head rotation, brow, eyes, and mouth. Within each batch, it reassigns the noise trajectories to reduce
their distance from the corresponding regional targets.

\paragraph{Autoregressive rollout.} Each 25-frame training target is divided into five successive chunks. For each
chunk, the model predicts its velocity and forms a one-step clean estimate
$\hat{\mathbf{m}}=\mathbf{m}_t+(1-t)\,\mathbf{v}_\theta(\mathbf{m}_t,t)$. A teacher-forcing curriculum determines
whether the rolling history is updated with the ground-truth chunk or this estimate. During the first epoch, all
predictions use ground-truth history. We then begin replacing ground-truth updates with model estimates, starting from
the final chunk. Each epoch moves this boundary one chunk earlier. After each update, we remove the oldest five frames
to keep the history at 75 frames. Ground-truth chunks are marked with $t=1$, while each estimate retains the sampled
timestep used to construct it. We add a learned embedding of this value to the past-motion tokens. This lets the decoder
distinguish ground truth from estimates formed at different timesteps. Conditioning on its own outputs during training
reduces exposure bias between training and autoregressive inference.

\paragraph{Guidance dropout.} We train the model for CFG~\cite{ho2022cfg} by randomly masking the condition paths in
Equation~\ref{eq:cond-residual}. We retain self audio, other audio, and the reference together for 45\% of samples and
drop all three for 15\%. The remaining samples retain one or two of these conditions. Past-motion history is masked
separately and retained for 75\% of samples. The model therefore learns to operate with any subset of the four
conditions. At inference, the past-only prediction forms the baseline, and self audio, other audio, and the reference
each add a guidance residual scaled independently for each motion region.

\paragraph{Optimization.} We set the objective coefficients $\lambda_{\mathrm{exp}}=0.1$,
$\lambda_{\mathrm{rot}}=\lambda_{\mathrm{3D}}=100$, and $\lambda_{\mathrm{VAD}}=0.01$. We optimize with
Adan~\cite{xie2022adan} using three weight-decay groups: 0.05 for most matrix weights, 0.01 for the AdaLN, timestep, and
reference projections, and zero for one-dimensional parameters, biases, self-attention gates, and quality embeddings.
The learning rate increases linearly over 5{,}000 warm-up steps to $3\times10^{-4}$, then follows a 200{,}000-step
cosine decay toward $10^{-5}$. We maintain an exponential moving average of the model weights with decay 0.995 and use
the averaged weights in the released checkpoint. The checkpoint was trained for 200{,}000 steps on a single NVIDIA GH200
in approximately 35 hours. Training uses float32 precision with TF32 matrix multiplication.

%% file: sections/hubert.tex
\subsection{Chunk-based Audio Encoder}
\label{sec:hubert}
The motion generation model is conditioned on two separate audio channels. During training, we extract features using
HuBERT~\cite{hsu2021hubert} on full-length audio. But, during real-time streaming inference, we can't do that, because
audio arrives incrementally, so the encoder must process audio in chunks. However, because HuBERT uses bidirectional
attention, directly encoding short chunks of audio produces features that significantly differ from their full-context
counterparts. To solve this, we build a chunk-based encoder through self-distillation of the original model without
changing its architecture. We set the input chunk to a duration of 525\,ms, semantically split into 120\,ms past, 200\,ms
present, and 205\,ms future windows, or 3, 5, and 5 video frames respectively. The extra 5\,ms of future audio is required
by HuBERT’s convolutional feature extractor to produce exactly 26 feature vectors at 50\,Hz that will be downsampled to 13
to match the number of video frames.

\paragraph{Feature drift.} We quantify the drift between fixed-window features and full-context features for the original HuBERT. For
a given present window, the features computed only on the window itself differ from the same window features computed on
full audio by 30--60\% in relative $L_1$ distance. Fidelity to the full-context features reaches a plateau at
approximately 2\,s of past audio and drops sharply with shorter past context. At least 400\,ms of future audio is
required to avoid severe degradation. This future lookahead exceeds the latency target, while processing two seconds of
past audio increases encoder work for each chunk.

\paragraph{Self-distillation.} Adapting a pretrained bidirectional speech encoder for chunk-based inference can require
architectural changes and continued self-supervised pretraining~\cite{fu2024wav2vecs}. We instead retain HuBERT's
architecture and fine-tune the encoder to match its own full-context features from a fixed input chunk. The teacher is
the frozen original encoder applied to the full-length audio. We initialize the student from the same weights and apply
it to an input of 525\,ms duration. Let $\mathbf{h}^{\text{chunk}}_i, \mathbf{h}^{\text{full}}_i \in \mathbb{R}^d$ denote
the student and teacher features at frame $i$, where $d=1024$. We minimize the mean-squared error over the $N=13$ frames
of the window:
\begin{equation}
  \mathcal{L}_{\text{distill}} = \frac{1}{Nd} \sum_{i=1}^{N}
  \left\lVert \mathbf{h}_i^{\text{chunk}} - \mathbf{h}_i^{\text{full}} \right\rVert_2^2 .
  \label{eq:hubert-mse}
\end{equation}
Under this window, self-distillation reduces the relative $L_1$ error from approximately 50\% to 12\%.

\paragraph{Training.} The encoder is trained on the English partition of Multilingual LibriSpeech~\cite{pratap2020mls}
and on the same dataset used to train the motion model. The convolutional feature extractor remains frozen, while the
feature projection and Transformer are optimized with AdamW (learning rate $10^{-3}$ and weight decay $10^{-4}$).
Training uses a Noam schedule with 20{,}000 warm-up steps, bfloat16 precision, and batch size 60, for approximately 200
epochs. For silence robustness, we replace 10\% of the input windows with silence or low-amplitude noise and supervise
them with a precomputed sequence of full-context silence features. A separate augmentation adds low-amplitude Gaussian
noise to a subset of speech windows.

%% file: sections/renderer.tex
\subsection{Renderer}
\label{sec:renderer}

\begin{figure}[t]
  \centering
  \includegraphics{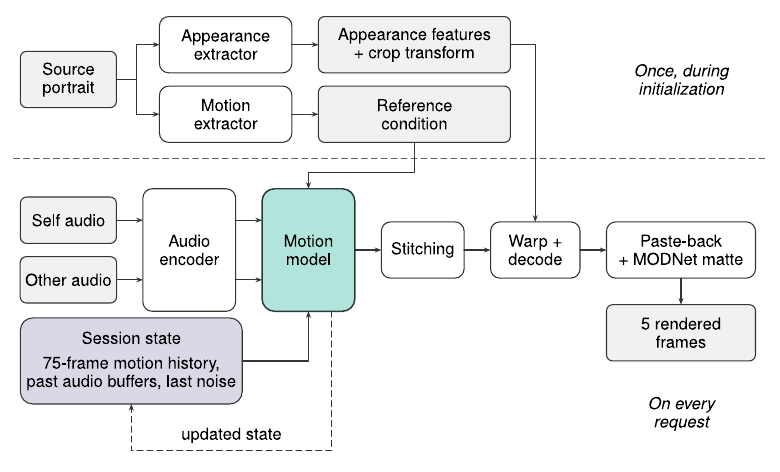}
  \caption{Overview of AVTR-1 inference.}
  \label{fig:inference}
\end{figure}

We call the inference-side component that serves the motion model the renderer. It registers the source portrait once,
then generates five frames at a time. Each request predicts a motion chunk, renders it with the cached LivePortrait
appearance~\cite{guo2024liveportrait}, and returns the updated state required by the next request.
Figure~\ref{fig:inference} shows the pipeline; Table~\ref{tab:inference-performance} reports its per-chunk request
processing duration and time to first frame across GPUs.

\paragraph{Session initialization.} Before generation, a face detector and two landmark models locate the face and
define its crop. The LivePortrait motion extractor predicts the source motion parameters, while the appearance extractor
produces the volumetric feature used for rendering. We cache these outputs together with the source frame, crop
transform, and pasteback mask. The source motion also provides the reference condition described in
Section~\ref{sec:motion}.

\paragraph{Audio context.}
The audio encoder and motion model use different context lengths. The encoder always receives a 525\,ms window, while the
motion model conditions on a longer window that includes a 75-frame audio history. For each motion chunk, the two audio
channels are batched into one audio encoder call. We keep its 5 present and 5 future features and discard the 3 past
outputs, because those are already cached from earlier calls. The runtime prepends 75 cached audio features as the past,
so the motion model's audio window spans 75 past, 5 present, and 5 future frames per stream. After generation, the 5
present features from both streams are added to the cache. The 5 future features condition the current chunk but are not
cached; the encoder recomputes them when they become present on the next step. This keeps the encoder's input fixed at a
13-frame window while preserving the 75-frame audio history the motion model requires.

\begin{figure}[t]
  \centering
  \includegraphics[width=0.8\linewidth]{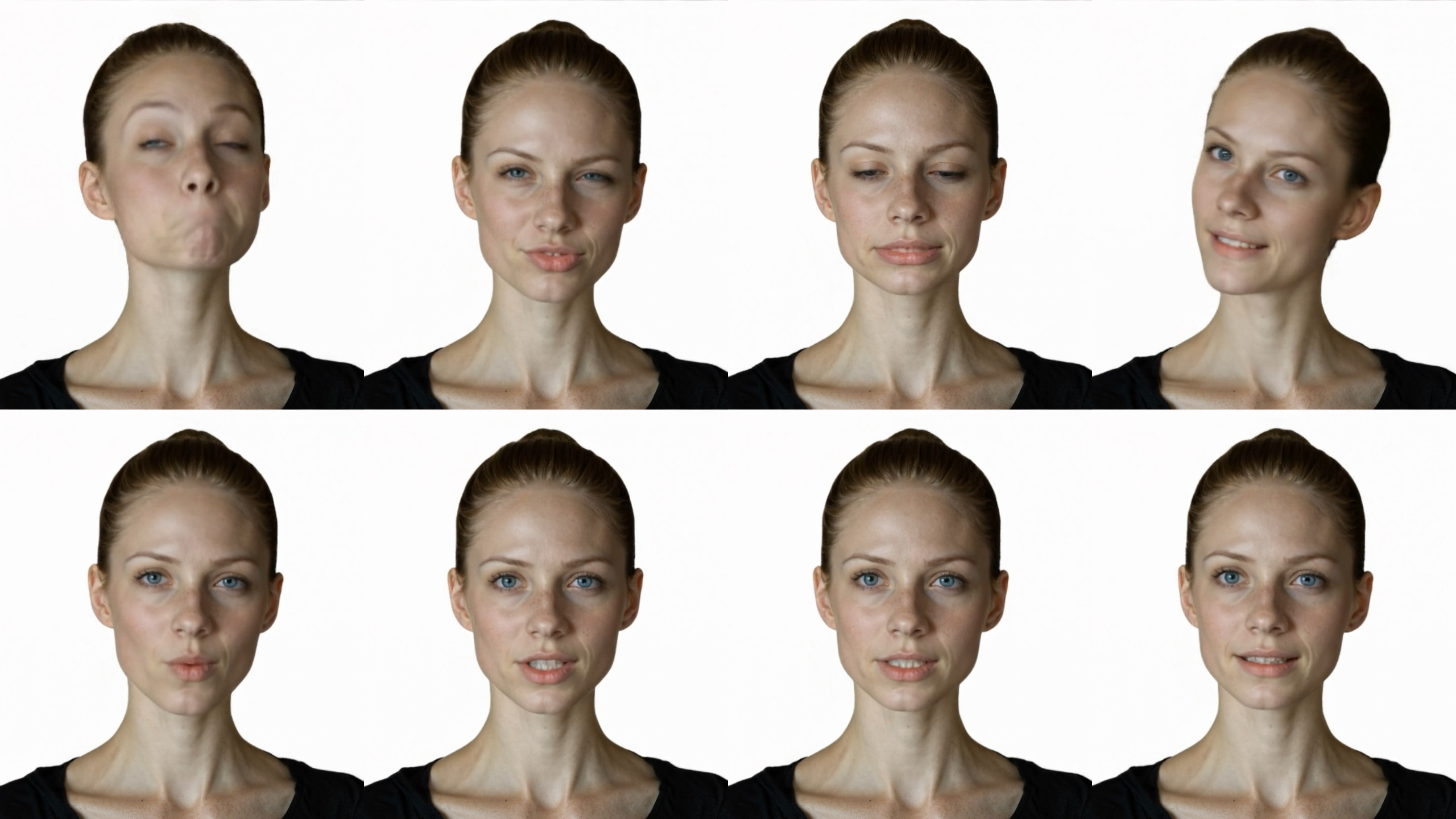}
  \caption{Qualitative effect of noise truncation. Representative frames generated without truncation (top) and with
  threshold $\tau=1.2$ (bottom). Truncation reduces extreme facial deformations and head rotations while preserving
  natural motion.}
  \label{fig:noise-truncation}
\end{figure}

\paragraph{Noise initialization and flow sampling.} At inference, we use the progressive noise described in
Section~\ref{sec:training} with correlation parameter $\alpha=2$. We additionally apply noise truncation with threshold
$\tau=1.2$, bounding every coordinate to $[-\tau,\tau]$. The truncation interval at each frame depends on the preceding
noise value, preserving the progressive correlation under this bound. In our qualitative examples, truncation suppresses
extreme expression deformations and head rotations observed without it (Figure~\ref{fig:noise-truncation}). We integrate
the learned velocity field using an Euler ODE solver with four model evaluations per chunk. The normalized prediction
then replaces the oldest five frames in the motion history.

\paragraph{Classifier-free guidance.} At inference, we apply classifier-free guidance in the regional latent spaces,
using the past-only representation as the baseline. At each solver step, the model evaluates four combinations in one
batch: past motion alone and past motion with self audio, other audio, or the reference condition. Let
$\boldsymbol{\ell}^{(q)}_{p}$, $\boldsymbol{\ell}^{(q)}_{\mathrm{self}}$, $\boldsymbol{\ell}^{(q)}_{\mathrm{other}}$,
and $\boldsymbol{\ell}^{(q)}_{\mathrm{ref}}$ denote the resulting latent representations for region $q \in
\{\mathrm{rotation}, \mathrm{brow}, \mathrm{eyes}, \mathrm{mouth}\}$. We compute
\begin{equation}
\label{eq:cfg}
\begin{aligned}
\widetilde{\boldsymbol{\ell}}^{(q)} = {}& \boldsymbol{\ell}^{(q)}_{p} + w^{(q)}_{\mathrm{self}}\left(\boldsymbol{\ell}^{(q)}_{\mathrm{self}} - \boldsymbol{\ell}^{(q)}_{p}\right) \\
&+ w^{(q)}_{\mathrm{other}}\left(\boldsymbol{\ell}^{(q)}_{\mathrm{other}} - \boldsymbol{\ell}^{(q)}_{p}\right) \\
&+ w^{(q)}_{\mathrm{ref}}\left(\boldsymbol{\ell}^{(q)}_{\mathrm{ref}} - \boldsymbol{\ell}^{(q)}_{p}\right).
\end{aligned}
\end{equation}
Here, the weights $w^{(q)}_{\mathrm{self}}$, $w^{(q)}_{\mathrm{other}}$, and $w^{(q)}_{\mathrm{ref}}$ scale the three
guidance residuals independently for each region. After regional RMS normalization, a linear projection produces the
velocity prediction for region $q$. The released runtime API uses one value per condition and applies it across all four
regions.

\paragraph{Rendering.} After flow sampling, we de-normalize the 42-dimensional motion and convert the axis-angle
rotation to a rotation matrix. We combine the predicted rotation and 39 expression coordinates with the source
parameters excluded from the training target (Section~\ref{sec:motion}), producing a set of driving keypoints for each
frame. The LivePortrait stitching network refines these keypoints based on the source keypoints. The renderer then uses
the refined keypoints to warp the cached appearance and decode the face crop. It pastes the crop back into the full
frame. MODNet~\cite{ke2022modnet} predicts an alpha matte for blending the rendered avatar with the selected background.

\paragraph{Session state.} We maintain a separate mutable state for each session. At the first request, its 75-frame
motion history repeats the source motion, its audio buffers start from zero, and it has no previous noise frame. After
every request, the state retains the motion history, 75 frames of dual-stream audio features, the 120\,ms of past audio
for both channels required by the audio encoder, and the final noise frame. These tensors occupy approximately 0.64\,MB.
The in-process runtime keeps the state on the GPU between chunks, while the released HTTP API serializes it between
requests.

\paragraph{TensorRT runtime.} The released runtime uses TensorRT engines for audio encoding, motion generation, and
rendering. The HuBERT streaming speech encoder engine runs once with batch size 2 to process the self and other audio
channels in parallel. The motion generator is split into separate encoder and decoder engines. Only the motion decoder
runs 4 times, for ODE integration. LivePortrait's Appearance Warping and Stitch engines process all generated frames in
parallel with an effective batch size of 5. Then LivePortrait's Face Decoder and MODNet process each frame sequentially
with batch size 1, for performance reasons. The source motion extractor runs once in fp32 during session initialization
in the ONNX runtime. During each request, the session state and intermediate tensors remain on the GPU and are reused
throughout flow sampling. The engines use module-specific batching and execute on a single CUDA stream without CPU
synchronization. All engines run in fp16, with normalization layers held in fp32.

\paragraph{Per-GPU performance.} Table~\ref{tab:inference-performance} reports the per-chunk request processing
duration, real-time factor, and time to first frame (TTFF) of a complete five-frame request across GPUs. We define the
real-time factor as the 200\,ms output duration divided by the request processing duration, and the TTFF as the summed
time of all engines up to the first returned frame.

\begin{table}[H]
\centering
\caption{Per-chunk latency, real-time throughput, and time to first frame (TTFF) across GPUs, measured offline on one
five-frame (200\,ms) chunk.}
\label{tab:inference-performance}
\begin{tabular}{@{}lccc@{}}
\toprule
GPU & TTFF & Request duration & Real-time factor \\
\midrule
NVIDIA L40S & 42\,ms & 71\,ms & $2.81\times$ \\
NVIDIA A100 & 54\,ms & 91\,ms & $2.20\times$ \\
NVIDIA L4 & --- & 202\,ms & $0.99\times$ \\
\midrule
NVIDIA RTX 4060 Ti & 99\,ms & 162\,ms & $1.24\times$ \\
NVIDIA RTX 3070 & 94\,ms & 180\,ms & $1.11\times$ \\
NVIDIA RTX 3060 Ti & --- & 207\,ms & $0.97\times$ \\
NVIDIA RTX 4060 & --- & 232\,ms & $0.86\times$ \\
\bottomrule
\end{tabular}
\end{table}

%% file: sections/streamer.tex
\subsection{Streamer}
\label{sec:streamer}

\begin{figure}[H]
  \centering
  \includegraphics{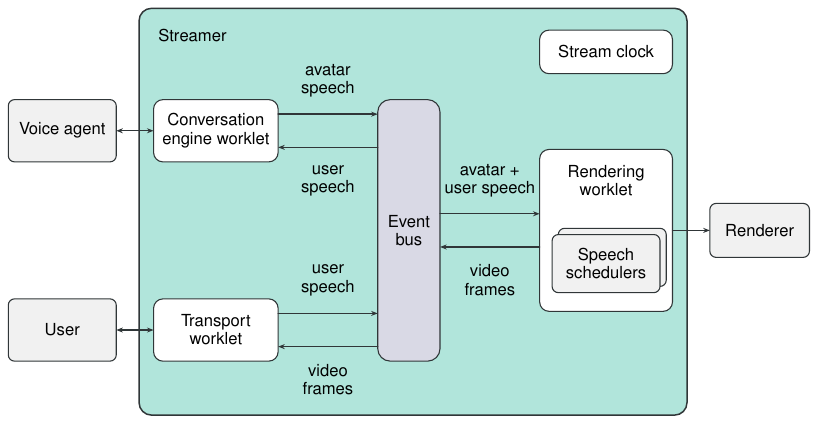}
  \caption{The streamer's worklet and event-bus architecture. A transport, a conversation engine, and a rendering
  worklet communicate only through a central event bus. The conversation engine worklet adapts a voice agent, the
  rendering worklet holds one speech scheduler per speech stream, avatar and user, and drives the renderer, returning
  fused audio-video frames. Worklets read the stream clock directly.}
  \label{fig:streamer}
\end{figure}

The motion model and, by extension, the renderer accept discrete fixed-size chunks of speech audio and produce a
fixed-size chunk of five frames, processing one chunk per request. Neither interprets user speech nor generates avatar
speech of its own. An interactive, video-call-like experience, by contrast, demands frames arriving at a constant rate,
an integration with a voice agent that listens and responds to the user, a bounded delay between the user finishing and
the avatar answering, and an avatar that stops speaking as soon as the user talks over it. The streamer is the runtime
that closes this gap. It connects the user, the voice agent, and a loop that continuously drives the renderer,
synchronizing their work against a shared timeline. It generates frames continuously rather than only while the avatar
speaks, so the model can exhibit the reactive listening it was trained for instead of freezing when the avatar is not
speaking. Figure~\ref{fig:streamer} shows the serving architecture.

\paragraph{Architecture.} The main functionality of the streamer is implemented by three independent components running
in parallel, which we call worklets: a transport, a conversation engine, and a rendering worklet. Together we call them
a media path. They are connected only by a type-routed event bus with an independent queue per consumer. It allows the
conversation engine and the transport to be replaced without changing the rendering loop. The bus also carries
control-plane concerns such as starting and terminating the stream, tracking user presence, and similar session-level
events. Control-plane worklets attach to the bus exactly as the media-path worklets do, and the media-path worklets
subscribe to their events.

\paragraph{Speech streams.} The streamer works with two independent speech streams: one for the user and one for the
avatar. The former is captured from the user's microphone and forwarded to the voice agent and to the model's other
audio channel. The latter is generated by the voice agent, sent to the model as the self audio channel, and then
delivered to the user together with the corresponding video frames. We represent both streams similarly. A speech stream
is a sequence of non-overlapping segments, and each segment is a start marker, one or more variable-sized audio
fragments, and an end marker. Fragments within a segment are expected to arrive no slower than real time under normal
conditions, which means that at any point in time the total duration of the fragments of the current segment that have
arrived so far must be equal to or greater than the time passed since the arrival of its start marker. The gap between
segments, on the other hand, can have an arbitrary duration.

\paragraph{Stream clock.}\label{par:stream-clock} Worklets synchronize their activity to a shared per-session stream
clock, which defines a session-local media timeline. The clock exposes two operations. A worklet can sleep until a
deadline on the timeline, or it can run an operation and have the clock measure how far past its deadline that operation
ran.

Measured overruns accumulate, and the clock subtracts the total from the time it reports, shifting the effective
timeline right by that amount. Worklets therefore read time as if no overrun has occurred. A deadline computed before an
overrun stays valid after it, and no worklet discovers itself late because of earlier lateness.

\paragraph{Transport worklet.} The transport worklet owns the user-facing communication and the jitter-free delivery of
outgoing media, and the released implementation drives a direct WebRTC peer. It publishes incoming user audio to the bus
as a speech stream, and it sends out the fused audio-video frames the rendering worklet produces. It releases each frame
at its media timestamp, maintaining a constant 25\,fps.

The transport segments the user speech stream by the availability of user microphone data rather than by speech content,
without any voice activity detection. A gap between segments therefore marks an interval in which no user audio was
available at all, whatever the cause, and not one in which the user wasn't speaking.

The transport starts the stream clock when it sends the first generated frame, which anchors the media timeline on first
delivery rather than on session setup. Connection negotiation, model initialization, and the first inference request
therefore cost nothing on the media timeline.

The transport also measures each frame against the time it should have been ready. When a frame arrives late, the clock
absorbs the overrun and shifts the timeline. The transport never skips frames or accelerates playback to compensate,
because either would corrupt the media. The delay is visible to the user as a pause of exactly its own duration. Every
generated sample is eventually played, and repeated delays leave a growing timeline offset rather than increase latency.

\paragraph{Conversation engine.} The avatar's speech comes from a voice agent, which in turn needs user speech to
respond to. A voice agent can be either an ASR-LLM-TTS pipeline or an end-to-end speech-to-speech model. It also decides
when the user has ended their turn and it should start responding, and when the user has started talking over it. The
conversation engine worklet is the adapter between the voice agent and the bus. It forwards user audio to the voice
agent, and it publishes the returned reply as a segment. It also forwards the voice agent's interruption requests to the
bus. The released implementation ships adapters for two voice agents: the OpenAI Realtime API, which is backed by
speech-to-speech models, and Cartesia Line agents, which are a managed ASR-LLM-TTS pipeline. The two share nothing but
the events they publish.

\paragraph{Speech scheduler.}\label{par:speech-scheduler} The speech streams produced by the conversation engine and the
transport have gaps between segments, and the fragments within a segment are arbitrarily sized and arrive at arbitrary
moments. The speech scheduler turns this irregular input into a continuous, gapless sequence of uniformly sized present-
and future-window pairs. The present and future window durations match the ones the model is trained for: 200 and 205\,ms
respectively. Below, completing a future window always means filling it to this full duration. The scheduler is not a
worklet but a component of the rendering worklet, which holds one instance per speech stream.

The scheduler places segment markers and fragments of the speech stream into its input queue as they arrive. On each
iteration it promotes a prefix of the future window to the present window unchanged and builds a new future-window
suffix that completes the window to its full duration, so the content of present window $k{+}1$ is decided on iteration
$k$. It completes the future window either by admitting markers and fragments from the input queue or by padding with
silence.

We call a segment active once its start marker and first fragment have been admitted into a future window and until its
end marker is admitted. Four rules govern admission:

\begin{itemize}
  \item \textbf{R1 (left padding).} When a segment's start marker and all of its queued fragments are admitted but their
  total duration is less than the window duration, the scheduler prepends silence, placing the fragments at the end of
  the window.
  \item \textbf{R2 (fragment splitting).} While a segment is active, if the window is still incomplete but the next
  queued fragment is longer than the still-required duration, the scheduler splits that fragment and places its suffix
  back at the head of the queue.
  \item \textbf{R3 (right padding at segment end).} When the scheduler admits a segment's end marker, it appends silence
  to complete the full window duration. When no segment is active, it fills the future window with silence entirely.
  \item \textbf{R4 (one boundary per window).} The scheduler never admits the next segment's start marker into a window
  in which another segment ends, which avoids a mid-window padding case.
\end{itemize}

Normally the scheduler does not block within an iteration. Occasionally, however, because of network latency or other
factors, fragments of an already active segment might arrive later than the window formation time, which leaves the
input queue empty while the future window is still short of its full duration. In this case the scheduler blocks the
caller until enough fragments have arrived to complete the window. It does not substitute silence for the missing
speech, for the same reason R1 does not right-pad within a segment.

When the conversation engine requests an interruption, the scheduler discards fragments and segments not yet admitted.
It also reports how much of the active segment had been admitted at the moment of interruption and which segments were
discarded entirely. The worklet publishes this information to the bus, where the conversation engine worklet picks it up
and forwards it to the voice agent. A voice agent can use it to remove the unheard part of the response from its
dialogue state if it generates speech faster than real time, making the state reflect what was heard rather than the
full response.

The scheduler also assigns media timestamps to segment start and end markers and to interruption and segment discard
events. The streamer delivers each event at its assigned media time rather than at the wall-clock moment the
conversation engine raised it.

\paragraph{Rendering worklet.} The rendering worklet generates frames continuously for the lifetime of the session. On
each iteration, it obtains present and future windows for both speech streams from the speech schedulers described above,
issues the inference request to the renderer immediately afterwards, and publishes the resulting five frames to the bus.
It pairs each frame with the corresponding 40\,ms of avatar speech audio before publication by slicing the present
window, so audio and video are aligned at the source rather than re-derived downstream from separately timestamped
streams. It also stamps each frame with its position on the media timeline, spaced by the 40\,ms frame duration.

Because frame generation takes time, each request must start some interval ahead of the moment its frames have to leave
the transport. We call that interval the lead time, $T_{\mathrm{lead}}$. If that moment has already passed, the request
goes out immediately rather than waiting.

The renderer streams frames one by one, and the worklet publishes each as soon as it is ready. The first frame arrives
after a longer delay, called the time to first frame (TTFF), because it follows the one-shot motion generation that
dominates a request, while the remaining four need only per-frame decoding and follow at shorter intervals. The whole
call must complete no slower than real time to keep the video jitter-free, the property the real-time factors in
Table~\ref{tab:inference-performance} report.

The lead time therefore only has to cover the TTFF. Later frames may arrive after the transport has already sent the
first frame, each still preceding its own output moment, and the hard bound is the last frame's output time. The system
therefore holds at most one chunk of rendered but unplayed video. Issuing the request earlier would add a delay that
every reply pays for, while issuing it later leaves no margin for variation in inference time. We set $T_{\mathrm{lead}}
= 100$\,ms, which covers the renderer's TTFF on our hardware. A more adaptive implementation could track the online TTFF
distribution and issue each request at, for instance, its 99th percentile ahead of the scheduled play time, fitting
$T_{\mathrm{lead}}$ to the serving hardware and its load.

The rendering loop is always active because the speech schedulers never block under normal conditions and can produce a
present and future window pair at the moment of the request. If either scheduler blocks mid-segment, the inference
request is delayed, so the frames are not available in the transport at the moment they have to be sent, which activates
the delay compensation of the stream clock.

The worklet also owns the per-session renderer state. The renderer holds none of it, so each request receives the state
the previous one returned.

Figure~\ref{fig:scheduling} illustrates the scheduler and the rendering worklet loop across several consecutive
iterations.

%% file: sections/latency.tex
\section{Latency}
\label{sec:latency}

Nothing above tells us how the avatar feels to interact with. The renderer's per-chunk performance and the scheduler's
admission rules describe how frames get produced, not how responsive the result is. Two latencies govern how an
interactive avatar feels in use: how long a reply takes to become audible, and how long ongoing speech keeps playing
after the user barges in. These are the established axes of turn-taking evaluation for voice agents, measured there as
response latency and stop latency~\cite{lin2025fullduplexbench15}, so they are the latencies we care about. Each of the
two decomposes into transport latency, the time to move a signal between the voice agent and the user, and internal
latency, the time to process it. Our system is a visual add-on that stays on the path of both: like a transport, it
forwards the voice agent's speech to the user with added video, so its own contribution appears in both latencies. Below
we show how latency forms across the scheduler's iterations and derive an analytical model for the streamer's
contribution to the two latencies, confirm the model experimentally, and then place that contribution inside a reference
end-to-end latency.

\subsection{Streamer Latency Model}
\label{sec:latency-model}

\begin{figure}[t!]
  \centering
  \includegraphics[width=\linewidth]{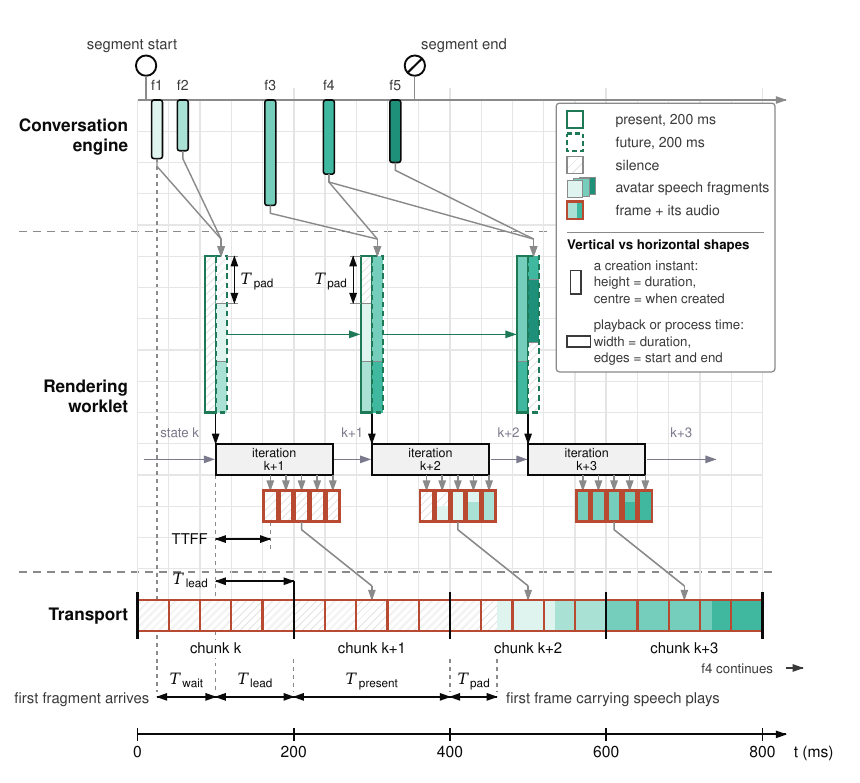}
  \caption{Streamer's scheduling and latency formation example. Three rendering-worklet iterations are shown. Each forms
  a present and a future window, issues one inference request, and returns five frames. For legibility the diagram draws
  the future window as 200\,ms rather than its actual 205\,ms. A tint identifies an arriving speech fragment and follows
  it into the windows that admit it, the frames that animate it, and the position in the transport output. Fragments
  arrive in a burst, faster than real time, so the speech for future window $k{+}3$ is already in the input queue when
  iteration $k{+}2$ begins. The segment is first admitted into the future window on iteration $k{+}1$ with silence
  prepended, is promoted to the present window on iteration $k{+}2$, and has its last fragment and end marker admitted
  into a future window on iteration $k{+}3$ with silence appended. Its fragment \texttt{f4} is only partially admitted
  on iteration $k{+}2$, and the remainder is admitted on iteration $k{+}3$. The first frame is ready before its chunk
  starts playing. The rest follow at gaps under 40\,ms, each before the moment it is due to play. The four terms
  composing the RLC, 435\,ms in this example, are marked along the bottom axis.}
  \label{fig:scheduling}
\end{figure}

Every admission rule of the speech scheduler is local: each decides only what enters one window on one iteration, and
none of them refers to latency. The streamer's contribution is therefore not a parameter of the design but a behavior
that emerges from composing those rules across iterations, which is why it has to be derived rather than stated.

We derive the bounds under three conditions the deployment must satisfy.

\textbf{Real-time speech stream.} The fragments of an active segment arrive no slower than real time: at any time after
a segment starts, the total duration of the fragments received so far is at least the elapsed time. This is the weakest
condition under which a live system can play a segment gaplessly, so any streaming TTS or speech-to-speech model must
satisfy it to be usable interactively.

\textbf{Observable speech segment markers.} The voice agent must expose the start and end of each contiguous speech
segment as discrete events. These are markers on the speech itself, not on conversational turns: a full-duplex model
such as Moshi~\cite{defossez2024moshi} has no turns, yet its speech still starts and stops, and only these speech
boundaries matter to the streamer. An agent that synthesizes audio continuously and emits no start marker fits our
representation only as a single session-long segment, so R1 never advances a reply into a partially filled window and
the RLC bound derived below does not apply.

\textbf{Non-blocking iterations.} Under real-time arrival the scheduler never blocks, because at window-formation time
the input queue holds enough of the active segment to fill the window. Transient network jitter can still leave the
queue short. The scheduler then blocks until enough fragments arrive rather than substituting silence, for the same
reason R1 does not right-pad within a segment. A blocking iteration delays its inference request, so the frames miss
their timestamps and the effect appears as a pause in playback rather than as added response latency, outside this
derivation.

We write $T$ for the response latency contribution (RLC): the time from the moment the conversation engine receives the
speech start marker and the first speech fragment of a reply until the first frame containing that reply's speech is
sent to the transport for playback. It decomposes into four terms. Because each iteration produces frames with a total
duration of $T_{\mathrm{present}} = 200$\,ms, one iteration starts in every $T_{\mathrm{present}}$ interval, so the
first fragment waits in the input queue for the start of the next iteration, $T_{\mathrm{wait}} \in [0, 200)$\,ms. That
iteration issues its request $T_{\mathrm{lead}} = 100$\,ms before its frames are scheduled to play. During the
iteration, the scheduler admits the fragment into the future window behind left silence padding, $T_{\mathrm{pad}} \in
[5, 205)$\,ms, because each iteration fills only a 200\,ms suffix of the future window, while a 5\,ms prefix carries
over from the previous window as non-replaceable constant padding. The fragment then waits one more iteration,
$T_{\mathrm{present}} = 200$\,ms, until it is promoted to the present window and its frames are generated. Under the
conditions above, the terms compose additively,
\[
  \mathbb{E}[T] = T_{\mathrm{lead}} + T_{\mathrm{present}} + \mathbb{E}[T_{\mathrm{wait}}] +
  \mathbb{E}[T_{\mathrm{pad}}],
\]
so $T$ has a floor of 305\,ms and stays strictly below 705\,ms. A realistic ceiling lies in the mid 600\,ms, because the
first speech fragment from TTS is typically at least a few tens of milliseconds. The mean cannot be established upfront,
however, because the distributions of $T_{\mathrm{wait}}$ and $T_{\mathrm{pad}}$ are deployment dependent. For example,
if the voice agent's TTS always produces a first fragment longer than 200\,ms, then $T_{\mathrm{pad}}$ stays at its
5\,ms floor. Likewise, if the voice agent operates at the same 200\,ms cadence but its iteration always starts 20\,ms
after ours, then $T_{\mathrm{wait}}$ is 180\,ms on every reply. None of the four terms refers to network distance, and
the renderer enters only through $T_{\mathrm{lead}}$, which has to cover the renderer's time to first frame (TTFF) and
nothing more, so the streamer's contribution is otherwise insensitive to the deployment. Figure~\ref{fig:scheduling}
marks the four terms component by component on its bottom axis.

The interruption latency contribution (ILC), which corresponds to stop latency, runs from the voice agent's interruption
signal to the end of avatar speech playback and decomposes into the same terms. The interruption signal still waits
$T_{\mathrm{wait}}$ for the next iteration, whose request is issued $T_{\mathrm{lead}}$ ahead of its frames, and the
$T_{\mathrm{present}}$ of audio admitted on the previous iteration is still promoted to the present window and plays in
full. Unlike the RLC, discarding unadmitted speech leaves the future window filled with silence by R3 rather than padded
around admitted speech, so $T_{\mathrm{pad}}$ falls to its constant 5\,ms floor. The ILC therefore shares the 305\,ms
floor and stays below 505\,ms.

\paragraph{Scheduler Trade-offs.}
Three of the four terms follow from scheduling choices that could be made differently. R1 accepts a bounded amount of
silence padding to keep latency low, rather than waiting for a full window of speech; waiting would remove the padding
but add a fixed 200~ms, worse in expectation for any first-fragment distribution with mean below 200~ms. Right-padding
instead of left-padding would also remove the padding, but would place a pause inside the speech and feed the motion
model out-of-domain data. A render-ahead depth of one chunk fixes the promotion term. Discarding only unadmitted speech
bounds interruption latency by exactly the committed present window; discarding committed media as well would lower the
floor but break the generator's expectation that the future window always becomes the present window. Only
$T_{\mathrm{lead}}$ is free.

\subsection{Measured Streamer Latency}
\label{sec:latency-measured}

We measure $T$ and its decomposition for the two commercial voice agent integrations shipped with our code, the OpenAI
Realtime API and Cartesia Line, reporting for each the first ten exchanges that carry both an avatar reply and an
interruption, so the measurements are paired and the sample size equal.

\newcommand{\latmean}[1]{\makebox[2.6em][r]{$#1$}} \newcommand{\latrange}[2]{\latmean{#1}\,{\scriptsize $#2$}}

\begin{table}[H]
  \centering
  \caption{Measured streamer latency for both voice agents, in milliseconds. Values are the mean followed by $[\min,
  \max]$, or a single value where constant. The predicted columns are the bounds derived above.}
  \label{tab:latency-agents}
  \small
  \setlength{\tabcolsep}{5.2pt}
  \begin{tabular}{@{}lllllll@{}}
    \toprule
     & \multicolumn{2}{c}{OpenAI Realtime API ($n = 10$)} & \multicolumn{2}{c}{Cartesia Line ($n = 10$)} & \multicolumn{2}{c}{\textbf{Predicted}} \\
    \cmidrule(lr){2-3} \cmidrule(lr){4-5} \cmidrule(lr){6-7}
    Term & RLC & ILC & RLC & ILC & RLC & ILC \\
    \midrule
    $T_{\mathrm{wait}}$   & \latrange{86.3}{[10, 187]}  & \latrange{58.8}{[6, 105]}   & \latrange{110.3}{[19, 193]} & \latrange{89.1}{[41, 169]}  & $[0, 200)$   & $[0, 200)$ \\
    $T_{\mathrm{pad}}$    & \latmean{5.0}                & \latmean{5.0}                & \latrange{45.0}{[5, 125]}   & \latmean{5.0}                & $[5, 205)$   & $5$ \\
    $T_{\mathrm{present}}$ & \latmean{200.0}             & \latmean{200.0}             & \latmean{200.0}             & \latmean{200.0}             & $200$        & $200$ \\
    $T_{\mathrm{lead}}$   & \latrange{99.1}{[98, 100]}  & \latrange{98.9}{[98, 100]}  & \latrange{98.9}{[97, 100]}  & \latrange{99.2}{[99, 100]}  & $100$        & $100$ \\
    \midrule
    Total                 & \latrange{390.4}{[315, 491]} & \latrange{362.7}{[310, 409]} & \latrange{454.2}{[426, 497]} & \latrange{393.3}{[345, 473]} & $[305, 705)$ & $[305, 505)$ \\
    \bottomrule
  \end{tabular}
\end{table}

Every measurement falls inside the predicted range, and in every record the terms sum to the total to within the 1\,ms
logging resolution, so the composition is additive rather than approximate under non-blocking conditions.
$T_{\mathrm{present}}$ holds at its configured 200\,ms for both agents. $T_{\mathrm{lead}}$ sits one to two milliseconds
below its configured 100\,ms, which is an implementation artifact rather than a voice-agent effect: the streamer runs
the rendering loop on a single thread, so it cannot issue the request at the exact $t_i - T_{\mathrm{lead}}$ instant and
instead starts the iteration as soon as it can after that moment, which shortens the effective lead by that small
amount. Neither term depends on the voice agent. The agents differ only in $T_{\mathrm{pad}}$, the term the model
predicts to be deployment dependent: the OpenAI fragments are large enough that a full future window is available at
admission, so only the 5\,ms constant prefix remains as padding and the RLC occupies only the lower half of its bound,
whereas Cartesia's smaller fragments raise $T_{\mathrm{pad}}$ above its 5\,ms floor in multiples of the 40\,ms frame
duration and, together with a slightly larger $T_{\mathrm{wait}}$, raise the mean RLC by 64\,ms. On interruption
$T_{\mathrm{pad}}$ falls to its 5\,ms floor for both, because discarding unadmitted speech fills the future window with
silence outright rather than padding around admitted speech.

\subsection{End-to-End Latency}
\label{sec:latency-e2e}

The RLC and ILC above are only the streamer's own contribution. The end-to-end response latency a user perceives, from
the end of their speech to the first frame of avatar speech, also includes the voice agent's own latency (speech
recognition, turn detection, and speech generation), the inbound and outbound transport between the user and the
streamer, and the communication between the streamer and the voice agent. Only turn detection and speech generation are
unique to interactive video avatar systems; the other terms are unavoidable for any voice-based system, with the
transport between the voice agent and the user replacing the one between the streamer and the user but keeping the same
nature. In production systems, voice agent latency is usually the dominant term, reaching
1--2~s~\cite{kramer2026voiceai,openbenchmarks2026voicelatency}, and transport latencies depend on conditions such as the
user and server locations. Covering all of this comprehensively is beyond our scope, so we report reference values for
the other terms from publicly available sources, for illustration only, and place the streamer's measured contribution
inside them.

\begin{table}[]
\centering
\caption{Reference end-to-end response latency budget for an avatar deployment, in milliseconds. The per-stage client
and voice-agent numbers are taken from the published voice-to-voice budget of~\cite{kramer2026voiceai} and grouped into
parts, and the bold part totals sum to the end-to-end figure. The two streamer $\leftrightarrow$ voice-agent network
hops and the streamer RLC are extras introduced by our system: the RLC is the measured mean for Cartesia Line
(Table~\ref{tab:latency-agents}), and the hops assume the streamer and the voice agent share a physical cloud region.}
\label{tab:serving-latency}
\newcommand{\budgetgroup}[2]{\textbf{#1} & \textbf{#2}} \newcommand{\budgetstage}[2]{\quad #1 & #2}
\begin{tabular}{@{}l r@{}}
\toprule
Stage & ms \\
\midrule
\budgetgroup{User $\rightarrow$ streamer (inbound)}{114} \\
\budgetstage{Microphone capture}{40} \\
\budgetstage{Opus encoding}{21} \\
\budgetstage{Network}{12} \\
\budgetstage{Jitter buffer}{40} \\
\budgetstage{Opus decoding}{1} \\
\addlinespace
\textbf{Streamer $\rightarrow$ voice agent network} & \textbf{5} \\
\addlinespace
\budgetgroup{Voice agent (end to end)}{1090} \\
\budgetstage{Transcription and endpointing}{300} \\
\budgetstage{LLM TTFB}{650} \\
\budgetstage{Sentence aggregation}{20} \\
\budgetstage{TTS TTFB}{120} \\
\addlinespace
\textbf{Voice agent $\rightarrow$ streamer network} & \textbf{5} \\
\addlinespace
\textbf{Streamer RLC $T$ (Cartesia Line)} & \textbf{454} \\
\addlinespace
\budgetgroup{Streamer $\rightarrow$ speaker output (outbound)}{89} \\
\budgetstage{Opus encoding}{21} \\
\budgetstage{Network}{12} \\
\budgetstage{Jitter buffer}{40} \\
\budgetstage{Opus decoding}{1} \\
\budgetstage{Speaker output}{15} \\
\midrule
\textbf{End to end} & \textbf{1757} \\
\bottomrule
\end{tabular}
\end{table}

Table~\ref{tab:serving-latency} places the streamer's contribution inside that budget. The voice agent dominates at
1090\,ms, the streamer is second at 454\,ms, and everything else together accounts for 213\,ms. The same source names
1500\,ms voice-to-voice as the target to aim for, which this budget exceeds: adding a video avatar to a voice stack that
already sits at 1293\,ms cannot reach that target, so a deployment aiming at it has to take the difference out of the
voice agent, where nearly two thirds of the budget is.

%% file: sections/evaluation.tex
\section{Evaluation}
\label{sec:evaluation}

We evaluate AVTR-1 along three axes: visual quality, speaking quality, and reactive listening.
Standard talking-head metrics cover the first two axes.
Several listening metrics compare facial motion between the two participants, but none isolates whether the
speaker's speech carries predictive information beyond the listener's own history and the speaker's motion.
We report these metrics alongside the Reference-Based Directed Granger Gain, which measures that additional
predictive contribution.

\paragraph{Protocol.}
We evaluate on 184 speaker--listener pairs from the conversational, improvised subset of the Seamless Interaction
(SI) test split~\cite{seamless2025}, totaling 11.90 hours at 30 fps.
We refer to this subset as SI-184.
For each pair, we retain the recorded speaker video, generate the listener, and compare it with the recorded listener.

\paragraph{R-DGG scored rows.}
For R-DGG, the common scored set after filtering contains 1,457 listening segments from 177 videos, totaling
3.46 hours at 25 fps.
Appendix~\ref{app:dataset} provides the full corpus selection and processing statistics.
We retain maximal intervals in which the listener remains silent and the speaker is active at least once.
We trim 0.2 seconds from listener-silence boundaries to reduce voice-activity errors and keep intervals with at
least four seconds remaining.
The listener motion is centered within each interval, and each scored row requires 100 preceding frames within
that interval, covering the complete four-second history.
For the common scored set, we retain only frames tracked successfully for ground truth and every evaluated
method.

\paragraph{Evaluated methods.}
We evaluate three dyadic systems, AvatarForcing*~\cite{ki2026avatarforcing}, AVTR-1, and
DyStream~\cite{chen2025dystream}, and four talking-head generators, Ditto~\cite{li2024ditto},
FLOAT~\cite{ki2024float}, SoulX Lite, and SoulX Pro~\cite{soulx}.
Every system receives the ground-truth audio track of the participant being generated, while the dyadic systems
also receive the paired participant's audio.
For the talking-head generators, the scored listening intervals correspond to portions of this track classified
as silent by voice-activity detection.
For AvatarForcing*, we extend the hardcoded RoPE range and increase the maximum generation length from 30 to
600 seconds so that it can process full conversations.
For R-DGG, we use $K=100$ circular shifts.
Appendix~\ref{app:temporal-offset-analysis} examines how the gain changes under temporal offsets of the speaker features.
The conventional metrics instead use all listener-silent frames selected by voice-activity detection, with
tracking validity handled separately for each method.
The metrics in Table~\ref{tab:motion-metrics} are computed from EMOCA features resampled to 25 fps, and
Appendix~\ref{app:liveportrait-metrics} reports their LivePortrait analogues.

\paragraph{Metrics.}
We use Fréchet Inception Distance (FID)~\cite{heusel2017fid} for frame-distribution quality, Fréchet Video
Distance (FVD)~\cite{unterthiner2018fvd} for video-distribution quality, and cosine similarity (CSIM) of ArcFace
features~\cite{deng2019arcface} for identity preservation.
We use SyncNet lip-sync error distance (LSE-D) and confidence (LSE-C)~\cite{chung2017syncnet} for audio-visual
synchronization.

For listener motion, Residual Pearson Correlation Coefficient (rPCC) measures the discrepancy between generated
and recorded speaker--listener motion correlations~\cite{tran2024dim}.
Paired Fréchet Distance (PFD) compares the generated and recorded joint distributions of speaker and listener
motion~\cite{ng2022learning}.
Shannon Index for Diversity (SID) measures the diversity of generated listener motion using K-means clustering,
while variance (Var) measures its variation over time~\cite{ng2022learning,ki2026avatarforcing}.
We compute these metrics from both EMOCA~\cite{danecek2022emoca} and
LivePortrait~\cite{guo2024liveportrait} features resampled to 25 fps.

\subsection{Reference-Based Directed Granger Gain}

A listener can follow the other participant's motion without responding to their speech.
The motion metrics above cannot distinguish this behavior from a speech-conditioned response.
To address this limitation, the Reference-Based Directed Granger Gain (R-DGG) measures how much the speaker's
speech improves prediction of the listener's current motion after accounting for the listener's own history and
the speaker's motion.
Following the Granger convention~\cite{granger1969}, the metric measures predictive dependence rather than
causal influence.
A shared event may affect both the speaker's speech and the listener's motion, producing a positive score without
the speech causing the listener's response.

\paragraph{Evaluation features.}
We represent each participant's motion with a 126-dimensional concatenation of EMOCA and LivePortrait features.
The listener representation at the current frame is the regression target, and its past values enter the
listener's own history.
We reduce the speaker representation by PCA to 32 dimensions before applying the reference projections.
Speech features for both participants come from layer 9 of HuBERT Base~\cite{hsu2021hubert} and are reduced by
PCA to 64 dimensions.
This evaluation encoder is separate from the streaming encoder described in Section~\ref{sec:hubert}.
All features are aligned on a common 25 fps grid.
Past motion and speech are sampled at 25 temporal offsets ranging from one to 100 frames.
For each offset, they are standardized to zero mean and unit variance using only rows where the offset remains
within the listening interval.

Let $v_d$ be the variance of coordinate $d$ in the ground-truth listener representation, and let
$v_{\mathrm{med}}$ be the median of these variances across all 126 coordinates.
We scale coordinate $d$ by
\begin{equation}
w_d=\left(v_d+v_{\mathrm{med}}\right)^{-1/2}.
\label{eq:target-scaling}
\end{equation}
The same weights are used for all evaluated methods.

\paragraph{Reference projections.}
Fitting separate speaker projections for each evaluated system would give each system its own notion of
responding, making the resulting scores incomparable.
Instead, we fit the reference projections on listening rows for which both EMOCA and LivePortrait successfully
extract the ground-truth listener representation.
Let $\mathbf{Y}^{\mathrm{GT}}$ contain the scaled ground-truth listener representation at these rows,
$\mathbf{X}^{\mathrm{own,GT}}$ its past motion and speech, $\mathbf{X}^{a}$ the speaker's past speech, and
$\mathbf{X}^{m}$ the speaker's past motion.
For $u\in\{a,m\}$, we fit
\begin{equation}
  (\mathbf{A}_{u},\mathbf{B}_{u})
  =\underset{\mathbf{A},\mathbf{B}}{\operatorname{arg\,min}}
  \left\lVert\mathbf{Y}^{\mathrm{GT}}-\mathbf{X}^{\mathrm{own,GT}}\mathbf{A}
  -\mathbf{X}^{u}\mathbf{B}\right\rVert_{F}^{2}
  +\lambda\left(\lVert\mathbf{A}\rVert_{F}^{2}+\lVert\mathbf{B}\rVert_{F}^{2}\right).
  \label{eq:granger-projection}
\end{equation}
Here, $\lVert\cdot\rVert_F$ denotes the Frobenius norm and $\lambda$ is the ridge regularization parameter,
obtained by multiplying $10^{-4}$ by the average squared norm of the regression input columns.
The fit with $u=a$ gives the speech projection $\mathbf{B}_{a}$, while $u=m$ gives the motion projection
$\mathbf{B}_{m}$.
Only these speaker projections are retained, and their outputs $\mathbf{X}^{a}\mathbf{B}_{a}$ and
$\mathbf{X}^{m}\mathbf{B}_{m}$ are used to score every system.
We estimate the projections by cross-fitting over identity-disjoint groups: conversations that share a participant
belong to the same group, which is scored using projections estimated on the remaining groups.

\paragraph{Scoring regressions.}
For each evaluated system, let $\mathbf{Y}$ contain its scaled listener representation and
$\mathbf{X}^{\mathrm{own}}$ its past motion and speech.
We define
\begin{equation}
\begin{aligned}
\mathbf{Z}_{m}&=[\mathbf{X}^{\mathrm{own}},\mathbf{X}^{m}\mathbf{B}_{m}],\\
\mathbf{Z}_{a,m}&=[\mathbf{X}^{\mathrm{own}},\mathbf{X}^{m}\mathbf{B}_{m},
\mathbf{X}^{a}\mathbf{B}_{a}].
\end{aligned}
\label{eq:granger-scoring-inputs}
\end{equation}
Two separate ridge regressions predict $\mathbf{Y}$ from these inputs on the common scored rows.
Let $\mathbf{R}_{m}$ and $\mathbf{R}_{a,m}$ denote their residual matrices.
The two regressions use the same listener history and projected speaker motion, while the second also includes
projected speaker speech.

\paragraph{Prediction gain.}
For each regression, we compute the total squared prediction error after scaling:
\begin{equation}
E_j=\lVert\mathbf{R}_{j}\rVert_F^2,
\qquad j\in\{m,(a,m)\}.
\label{eq:granger-residual-error}
\end{equation}
The prediction gain from adding projected speaker speech is
\begin{equation}
g_{a\mid m}=\log\frac{E_m}{E_{a,m}}.
\label{eq:granger-aligned-gain}
\end{equation}
We use the natural logarithm, so both $g_{a\mid m}$ and its shift-corrected form are measured in natural-log
units (nats).
A positive value indicates that projected speaker speech reduces prediction error beyond what listener history
and projected speaker motion already predict.

\paragraph{Temporal-shift correction.}
Because the scoring regressions are fitted and evaluated on the same temporally correlated rows, regression
flexibility can produce a positive gain even when the speaker features contain no information about listener
motion.
We therefore hold the reference projections fixed and refit the two scoring regressions for $K$ circular shifts
of the speaker features.
For each shift, every speaker--listener pair receives an independently sampled offset applied jointly to the
speaker's motion and speech.
Offsets shorter than four seconds are excluded.
The reported R-DGG score is
\begin{equation}
G_{a\mid m}=g_{a\mid m}-\frac{1}{K}\sum_{k=1}^{K}g_{a\mid m}^{(k)}.
\label{eq:granger-shift-corrected}
\end{equation}

\paragraph{Interpretation.}
A positive $G_{a\mid m}$ means that speaker speech is more predictive at its original timing than when temporally
shifted.
R-DGG measures this predictive dependence, not listening quality.
Generated motion can differ substantially from the recorded response and still score highly, and an overly
deterministic model may score above ground truth.
A value near zero does not establish the absence of a response.
The reference projections are linear and cover delays of up to four seconds, so they may miss responses at longer
or variable delays or responses that do not follow a consistent linear pattern.
Because the feature transforms, reference projections, target scaling, and shift correction are corpus-specific,
scores should be compared only among systems evaluated on the same corpus.

\paragraph{Uncertainty.}
Frames from the same conversation are correlated, and multiple conversations may contain the same participant.
We therefore estimate uncertainty by resampling participants rather than individual frames.
Each of 20,000 bootstrap draws samples participants with replacement and recomputes R-DGG from their prediction
errors.
We report the bootstrap median, the 2.5th--97.5th percentile interval, and
$p=\widehat{\Pr}(G_{a\mid m}\leq0)$.

\paragraph{Validation controls.}
We validate R-DGG with one positive case and two negative controls.
The ground-truth listener, scored with reference projections estimated from other identity groups, is the
positive case and should have an interval above zero.
The talking-head generators do not receive the paired participant's inputs, so their intervals should include
zero.
For the second negative control, denoted GT~$\times$~other, we pair each speaker's motion and speech with
ground-truth listener motion from a different identity and interaction.
We apply a reproducible random temporal offset and repeat or crop the listener motion to match the target
duration.
Rows whose four-second history includes the transition from the end of the source sequence to its beginning are
excluded.
This preserves natural listener motion while removing the actual speaker--listener pairing, and its interval
should include zero.
We interpret R-DGG on a corpus only when the ground-truth interval remains above zero and both types of negative
control include zero.

\subsection{Quantitative Results}

Quantitative comparisons are presented in Tables~\ref{tab:eval-results}, \ref{tab:motion-metrics}, and
\ref{tab:granger-results}.
AVTR-1 achieves the strongest overall performance among the compared dyadic systems, leading all visual-quality
metrics and most conventional listening-motion metrics while remaining competitive in lip synchronization.

\begin{table}[ht]
  \centering
  \small
  \setlength{\tabcolsep}{3pt}
  \caption{Quantitative comparison with competing methods on the SI-184 subset. Gray marks the best value among
  the dyadic systems in both panels. For listening metrics without an arrow, values closer to ground truth are better.
  The pose components of PFD and variance are reported in units of $10^{-2}$.
  AvatarForcing* denotes the evaluation configuration described in the protocol.}
  \label{tab:eval-results}
  \begin{subtable}{\linewidth}
  \centering
  \caption{Visual quality and lip synchronization.}
  \label{tab:eval-visual}
  \begin{tabular}{lcrrrrr}
    \toprule
    & & \multicolumn{3}{c}{Visual quality}
    & \multicolumn{2}{c}{Lip synchronization} \\
    \cmidrule(lr){3-5}
    \cmidrule(lr){6-7}
    Method
    & Paired audio
    & FID $\downarrow$
    & FVD $\downarrow$
    & CSIM $\uparrow$
    & LSE-D $\downarrow$
    & LSE-C $\uparrow$ \\
    \midrule
    SoulX Lite
    & No & 12.1 & 89.3 & 0.91 & 6.74 & 3.38 \\
    SoulX Pro
    & No & 9.4 & 53.6 & 0.93 & 7.35 & 3.09 \\
    Ditto
    & No & 16.0 & 113.6 & 0.95 & 7.12 & 3.11 \\
    FLOAT
    & No & 11.7 & 83.2 & 0.90 & 6.99 & 2.93 \\
    \midrule
    AvatarForcing*
    & Yes & 14.4 & 85.8 & 0.77 & 7.32 & 2.41 \\
    DyStream
    & Yes & 43.1 & 119.4 & 0.87 & \cellcolor{significantgray}6.57 & 3.21 \\
    AVTR-1 (Ours)
    & Yes & \cellcolor{significantgray}14.3 & \cellcolor{significantgray}76.8
    & \cellcolor{significantgray}0.94 & 7.08 & \cellcolor{significantgray}3.28 \\
    \bottomrule
  \end{tabular}
  \end{subtable}

  \medskip
  \begin{subtable}{\linewidth}
  \centering
  \caption{Conventional listening-motion metrics among dyadic systems.}
  \label{tab:eval-dyadic-listening}
  \begin{tabular}{lrrrrrrrr}
    \toprule
      & \multicolumn{2}{c}{rPCC $\downarrow$} & \multicolumn{2}{c}{PFD $\downarrow$}
      & \multicolumn{2}{c}{SID} & \multicolumn{2}{c}{Var} \\
    \cmidrule(lr){2-3}\cmidrule(lr){4-5}\cmidrule(lr){6-7}\cmidrule(lr){8-9}
    Method & Exp & Pose & Exp & Pose & Exp & Pose & Exp & Pose \\
    \midrule
    AvatarForcing* & 0.109 & 0.163 & 37.51 & 7.560 & 4.749 & \cellcolor{significantgray}3.848
      & 1.213 & 2.099 \\
    DyStream & 0.128 & 0.141 & 40.87 & 7.598 & 4.497 & 3.606
      & 1.164 & \cellcolor{significantgray}1.731 \\
    AVTR-1 (Ours) & \cellcolor{significantgray}0.083 & \cellcolor{significantgray}0.140
      & \cellcolor{significantgray}25.98 & \cellcolor{significantgray}6.417
      & \cellcolor{significantgray}4.970 & 3.254 & \cellcolor{significantgray}1.313 & 0.930 \\
    \midrule
    Ground truth & 0.000 & 0.000 & 0.000 & 0.000 & 5.231 & 4.001 & 1.463 & 1.789 \\
    \bottomrule
  \end{tabular}
  \end{subtable}
\end{table}

\begin{table}[ht]
  \centering
  \small
  \setlength{\tabcolsep}{3.0pt}
  \caption{Comparison using conventional motion-based listening metrics on the VAD-selected listener-silent
  frames of the SI-184 subset.
  This extends Table~\ref{tab:eval-dyadic-listening} with the talking-head generators.
  For metrics without an arrow, values closer to ground truth are better.
  The pose components of PFD and variance are reported in units of $10^{-2}$.
  Dark and light gray mark the best and second-best methods, respectively.}
  \label{tab:motion-metrics}
  \begin{tabular}{lccccccccc}
    \toprule
      & Paired audio & \multicolumn{2}{c}{rPCC $\downarrow$} & \multicolumn{2}{c}{PFD $\downarrow$}
      & \multicolumn{2}{c}{SID} & \multicolumn{2}{c}{Var} \\
    \cmidrule(lr){3-4}\cmidrule(lr){5-6}\cmidrule(lr){7-8}\cmidrule(lr){9-10}
    Method & & Exp & Pose & Exp & Pose & Exp & Pose & Exp & Pose \\
    \midrule
    AvatarForcing* & Yes & 0.109 & 0.163 & 37.51 & 7.560 & 4.749 & \cellcolor{bestgray}3.848
      & 1.213 & \cellcolor{secondgray}2.099 \\
    AVTR-1 & Yes & 0.083 & 0.140 & 25.98 & 6.417 & 4.970 & 3.254 & 1.313 & 0.930 \\
    DyStream & Yes & 0.128 & 0.141 & 40.87 & 7.598 & 4.497 & \cellcolor{secondgray}3.606
      & 1.164 & \cellcolor{bestgray}1.731 \\
    Ditto & No & 0.092 & 0.168 & 35.02 & 6.790 & \cellcolor{secondgray}5.006 & 3.253
      & 1.604 & 1.386 \\
    FLOAT & No & 0.117 & \cellcolor{secondgray}0.132 & 34.09 & \cellcolor{bestgray}4.753
      & 4.322 & 2.961 & 1.007 & 0.843 \\
    SoulX Lite & No & \cellcolor{secondgray}0.077 & 0.149 & \cellcolor{secondgray}25.67 & 6.501
      & 4.935 & 3.309 & \cellcolor{secondgray}1.335 & 1.013 \\
    SoulX Pro & No & \cellcolor{bestgray}0.071 & \cellcolor{bestgray}0.129
      & \cellcolor{bestgray}24.26 & \cellcolor{secondgray}5.418 & \cellcolor{bestgray}5.060
      & 3.354 & \cellcolor{bestgray}1.386 & 1.001 \\
    \midrule
    Ground truth & & 0.000 & 0.000 & 0.000 & 0.000 & 5.231 & 4.001 & 1.463 & 1.789 \\
    \bottomrule
  \end{tabular}
\end{table}

\paragraph{Limitations of motion-based metrics.}
Table~\ref{tab:motion-metrics} illustrates why conventional motion-based metrics do not establish dependence on
paired audio.
Talking-head generators occupy both highlighted ranks for every rPCC and PFD component and for expression SID
and variance.
Dyadic methods occupy both ranks only for pose SID and variance.
Appendix~\ref{app:liveportrait-metrics} repeats the comparison using LivePortrait features and again places
talking-head generators among the top-ranked methods.

\begin{table}[ht]
  \centering
  \small
  \setlength{\tabcolsep}{3.5pt}
  \caption{Comparison using R-DGG on the SI-184 subset.
  Values are reported in units of $10^{-4}$ nats.
  The final column reports $p=\widehat{\Pr}(G_{a\mid m}\leq0)$.
  Gray marks estimates whose intervals exclude zero.}
  \label{tab:granger-results}
  \begin{tabular}{lccrrrc}
    \toprule
      & Paired audio & R-DGG & \multicolumn{3}{c}{Bootstrap percentile} & $p$ \\
    \cmidrule(lr){4-6}
    Method & & & 2.5\% & 50\% & 97.5\% & \\
    \midrule
    AvatarForcing* & Yes & \cellcolor{significantgray}0.32 & 0.07 & 0.32 & 0.56 & 0.008 \\
    AVTR-1 & Yes & \cellcolor{significantgray}0.57 & 0.19 & 0.57 & 0.98 & 0.002 \\
    DyStream & Yes & \cellcolor{significantgray}0.48 & 0.06 & 0.49 & 0.93 & 0.011 \\
    Ditto & No & 0.06 & -0.20 & 0.06 & 0.31 & 0.335 \\
    FLOAT & No & 0.01 & -0.33 & 0.00 & 0.42 & 0.491 \\
    SoulX Lite & No & -0.22 & -0.54 & -0.22 & 0.21 & 0.857 \\
    SoulX Pro & No & 0.18 & -0.34 & 0.17 & 0.58 & 0.239 \\
    \midrule
    Ground truth & & \cellcolor{significantgray}0.72 & 0.32 & 0.72 & 1.12 & $<0.001$ \\
    GT $\times$ other & & -0.11 & -0.56 & -0.11 & 0.45 & 0.659 \\
    \bottomrule
  \end{tabular}
\end{table}

\FloatBarrier
\paragraph{R-DGG results.}
Table~\ref{tab:granger-results} shows that R-DGG passes its validation check: the ground-truth interval remains
above zero, while the interval for GT~$\times$~other includes zero.
All three dyadic systems have intervals above zero, while the intervals for every talking-head generator include
zero.
The intervals of the three dyadic systems overlap, so these results do not support a reliable ranking among them.

Taken together, these results show that AVTR-1 combines visual and speaking quality comparable to the evaluated
systems with listening motion that is predictively related to the paired speaker's speech.

%% file: sections/conclusion.tex
\section{Conclusion}
\label{sec:conclusion}

We present AVTR-1, a complete open stack for live interactive avatars.
It comprises a compact autoregressive flow-matching motion model conditioned on two audio channels and an inference system that turns chunk-based generation into a real-time conversation between a user and an avatar.
The evaluation demonstrates AVTR-1 is the strongest overall performer among the evaluated dyadic systems.

Our analysis shows that the system's contribution to response and interruption latency depends not only on model generation latency but also on how it forms audio windows and schedules renderer requests.
In our system, speech playback begins only after iteration wait, left padding, present-window promotion, and rendering lead, while an interruption takes effect after the already scheduled present window finishes playing.
Measurements with two commercial voice agents fall within the bounds derived from these scheduling rules.

Conventional motion-based metrics do not consistently separate systems with access to paired speech from those without it.
R-DGG complements these metrics by measuring whether speaker speech provides additional information for predicting listener motion beyond listener history and speaker motion.
It indicates speech dependence for ground-truth listener motion and all evaluated dyadic systems, but not for negative controls.
This score should not be interpreted as listening quality or causal influence.
We release the model weights, renderer, and streamer under component-specific licenses.

%% file: sections/appendix.tex
\section{Qualitative Comparison}
\label{sec:qualitative-comparison}

\begin{figure}[H]
  \centering
  \includegraphics[height=0.76\textheight]{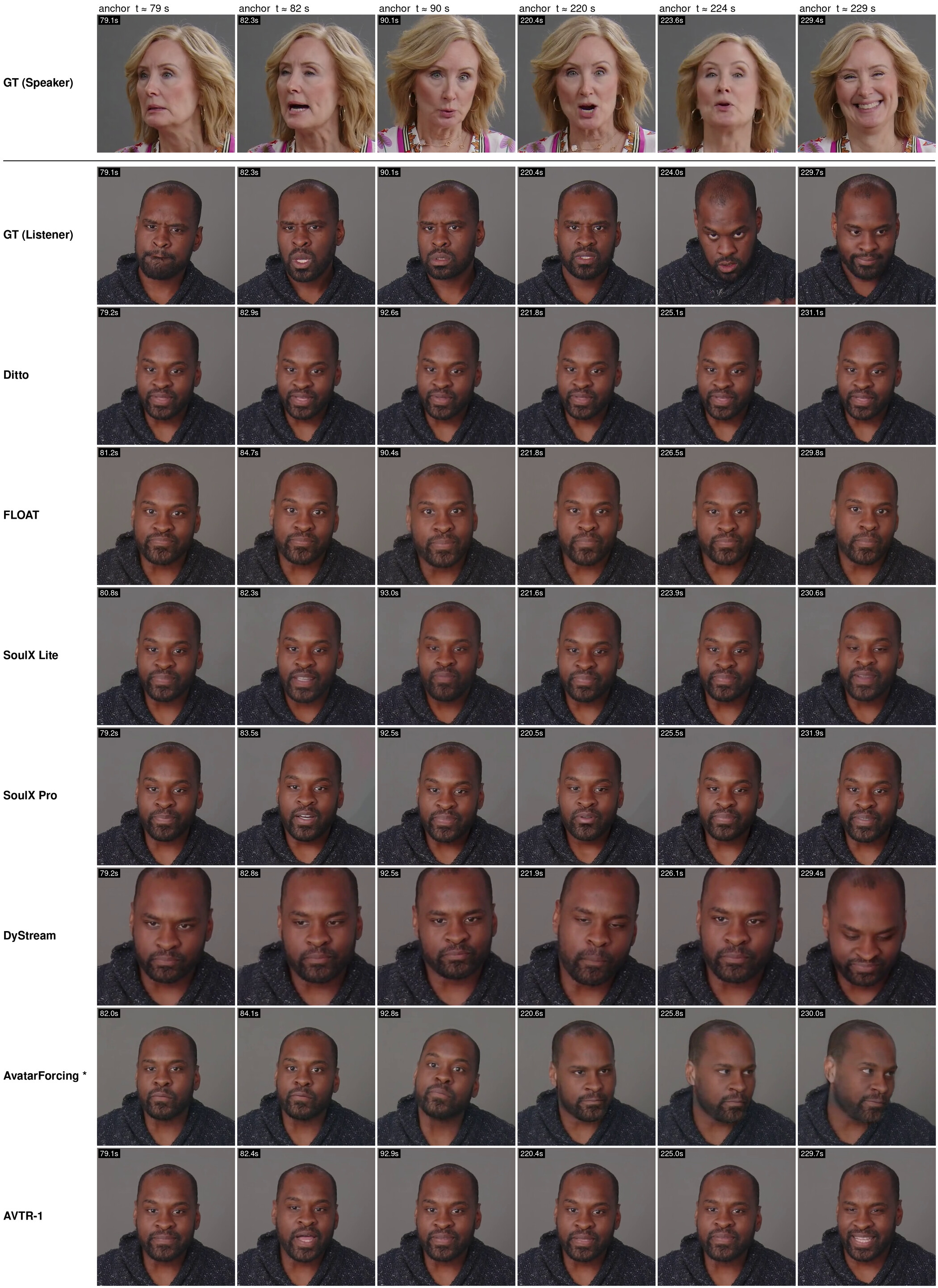}
  \caption{Qualitative comparison of reactive listening on the Seamless Interaction test set. The first two rows
  show the recorded speaker and listener, followed by frames generated for the listener by each evaluated system.
  Columns are anchored to the speaker times shown above the first row; inset labels report the selected timestamp
  for each video.}
  \label{fig:qualitative-comparison-1}
\end{figure}

\clearpage

\begin{figure}[H]
  \centering
  \includegraphics[height=0.84\textheight]{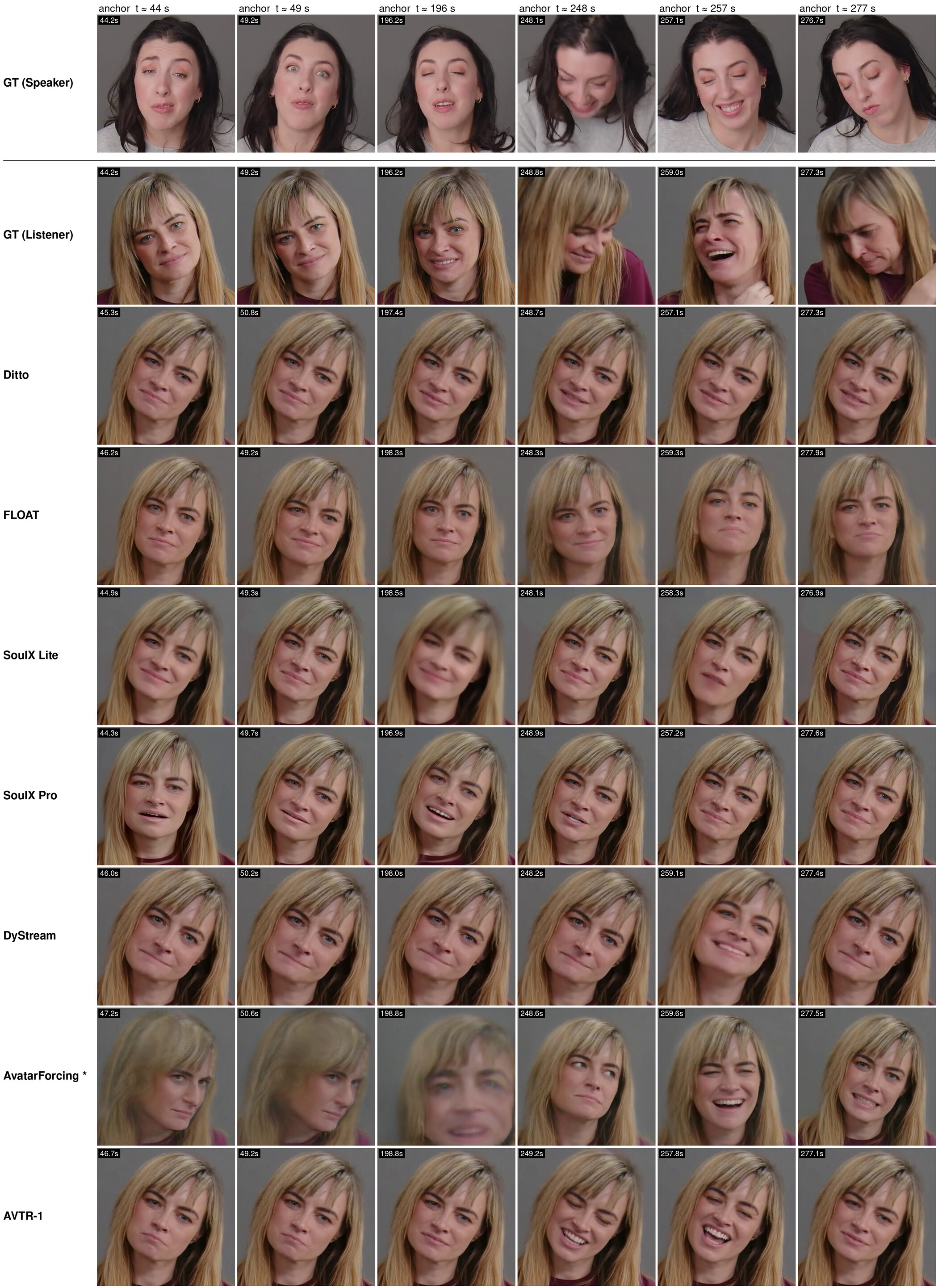}
  \caption{Additional qualitative comparison of reactive listening on the Seamless Interaction test set. The first
  two rows show the recorded speaker and listener, followed by frames generated for the listener by each evaluated
  system. Columns are anchored to the speaker times shown above the first row; inset labels report the selected
  timestamp for each video.}
  \label{fig:qualitative-comparison-2}
\end{figure}

\section{Evaluation Dataset Processing}
\label{app:dataset}

The dataset is identified by the following Seamless Interaction configuration:
\begin{center}
\begin{tabular}{ll}
  \toprule
  Configuration field & Value \\
  \midrule
  \texttt{vendor} & \texttt{V00} \\
  \texttt{interaction\_type} & \texttt{ipc\_conversation} \\
  \texttt{label} & \texttt{improvised} \\
  \texttt{split} & \texttt{test} \\
  \bottomrule
\end{tabular}
\end{center}
The broader \texttt{V00}/\texttt{improvised}/\texttt{test} selection contains 186 videos: 184 conversational
videos and two \texttt{grounded\_gesture} videos from interaction
\texttt{V00\_S2050\_I00001256}, corresponding to participants \texttt{P1307A} and \texttt{P1308A}.
We exclude the latter through the \texttt{interaction\_type} filter because they depict a grounded physical
performance rather than a conversation, for which speech-conditioned listening is not well-defined.

Before listening-segment selection, the 184 videos comprise 1,284,704 frames, or 42,823 seconds (11.90 hours),
at 30 fps.
Individual videos range from 138 to 346 seconds, with a 25th percentile, median, mean, and 75th percentile of
212, 233, 233, and 257 seconds, respectively.
Listening segments are derived from ground-truth voice activity, retaining intervals in which the listener is
silent and the speaker is active at least once.
We additionally require successful face tracking for ground truth and every evaluated output, trim 0.2 seconds
from silence boundaries, discard segments shorter than four seconds, and mask temporal offsets that precede the
start of a segment.
After these filters, 177 videos contribute 1,457 segments and 311,776 frames on the 25 fps evaluation grid,
equivalent to 3.46 hours of listening motion.
The mean segment duration is 8.6 seconds.
Seven videos contain no valid listening segment of at least four seconds.

\section{Additional Evaluation}
\label{app:additional-evaluation}

Tables~\ref{tab:eval-dyadic-listening} and~\ref{tab:motion-metrics} report conventional motion-based metrics
computed from EMOCA features on the VAD-masked listening portion of the SI-184 subset.
This appendix extends the evaluation across the motion representation and the set of evaluated frames.
AvatarForcing* denotes the evaluation configuration described in Section~\ref{sec:evaluation}.

\subsection{VAD-Masked Evaluation with LivePortrait Features}
\label{app:liveportrait-metrics}

Table~\ref{tab:motion-metrics-liveportrait} repeats the VAD-masked comparison using LivePortrait features.
This comparison tests whether the pattern observed with EMOCA depends on the motion representation.

\begin{table}[ht]
  \centering
  \small
  \setlength{\tabcolsep}{3.0pt}
  \caption{Comparison using conventional motion-based listening metrics computed from LivePortrait features on
  the VAD-masked listening portion of the SI-184 subset.
  For metrics without an arrow, values closer to ground truth are better.
  Expression PFD and variance are reported in units of $10^{-3}$ and $10^{-5}$, respectively.
  Their rotation components are reported in units of $10^{-2}$.
  Dark and light gray mark the best and second-best methods, respectively.}
  \label{tab:motion-metrics-liveportrait}
  \begin{tabular}{lccccccccc}
    \toprule
      & Paired audio & \multicolumn{2}{c}{rPCC $\downarrow$} & \multicolumn{2}{c}{PFD $\downarrow$}
      & \multicolumn{2}{c}{SID} & \multicolumn{2}{c}{Var} \\
    \cmidrule(lr){3-4}\cmidrule(lr){5-6}\cmidrule(lr){7-8}\cmidrule(lr){9-10}
    Method & & Exp & Rot & Exp & Rot & Exp & Rot & Exp & Rot \\
    \midrule
    AvatarForcing* & Yes & 0.138 & 0.175 & 1.242 & 5.784 & 4.802 & \cellcolor{bestgray}3.813
      & 5.200 & 3.352 \\
    AVTR-1 & Yes & 0.112 & \cellcolor{secondgray}0.169 & \cellcolor{secondgray}0.671 & 3.967
      & 4.760 & 3.152 & 3.300 & 0.920 \\
    DyStream & Yes & 0.151 & 0.178 & 1.335 & 4.392 & 4.421 & \cellcolor{secondgray}3.436
      & \cellcolor{secondgray}4.300 & \cellcolor{bestgray}1.239 \\
    Ditto & No & 0.109 & 0.174 & 0.882 & 4.106 & \cellcolor{secondgray}4.930 & 3.010
      & \cellcolor{bestgray}3.800 & 1.002 \\
    FLOAT & No & 0.127 & \cellcolor{bestgray}0.152 & 0.878 & \cellcolor{bestgray}3.240
      & 4.488 & 3.061 & \cellcolor{secondgray}4.300 & 0.852 \\
    SoulX Lite & No & \cellcolor{secondgray}0.098 & 0.188 & 0.685 & 4.018
      & 4.914 & 3.140 & 3.600 & 0.857 \\
    SoulX Pro & No & \cellcolor{bestgray}0.094 & 0.170 & \cellcolor{bestgray}0.633
      & \cellcolor{secondgray}3.332 & \cellcolor{bestgray}4.997 & 3.271
      & \cellcolor{secondgray}3.700 & \cellcolor{secondgray}1.013 \\
    \midrule
    Ground truth & & 0.000 & 0.000 & 0.000 & 0.000 & 5.171 & 3.745 & 4.000 & 1.900 \\
    \bottomrule
  \end{tabular}
\end{table}

\FloatBarrier

\subsection{Full-Frame Evaluation}

The full protocol uses all successfully tracked frames, whereas the VAD-masked protocol retains only frames on
which the listener is classified as silent.
Tables~\ref{tab:motion-metrics-emoca-full} and~\ref{tab:motion-metrics-liveportrait-full} report the corresponding
full-frame results using EMOCA and LivePortrait features.

\begin{table}[ht]
  \centering
  \small
  \setlength{\tabcolsep}{3pt}
  \caption{Conventional motion-based metrics computed from EMOCA features on all frames of the SI-184 subset.
  For metrics without an arrow, values closer to ground truth are better.
  The pose components of PFD and variance are reported in units of $10^{-2}$.
  Gray marks the best value among dyadic systems.}
  \label{tab:motion-metrics-emoca-full}
  \begin{tabular}{lcrrrrrrrr}
    \toprule
      & & \multicolumn{2}{c}{rPCC $\downarrow$} & \multicolumn{2}{c}{PFD $\downarrow$}
      & \multicolumn{2}{c}{SID} & \multicolumn{2}{c}{Var} \\
    \cmidrule(lr){3-4}\cmidrule(lr){5-6}\cmidrule(lr){7-8}\cmidrule(lr){9-10}
    Method & Paired audio & Exp & Pose & Exp & Pose & Exp & Pose & Exp & Pose \\
    \midrule
    SoulX Lite & No & 0.061 & 0.148 & 19.94 & 5.988
      & 5.065 & 3.419 & 1.411 & 0.975 \\
    SoulX Pro & No & 0.056 & 0.125 & 18.75 & 4.751
      & 5.103 & 3.499 & 1.441 & 0.982 \\
    Ditto & No & 0.074 & 0.180 & 28.82 & 6.645
      & 4.953 & 3.459 & 1.671 & 1.341 \\
    FLOAT & No & 0.077 & 0.128 & 23.71 & 3.953
      & 4.816 & 3.220 & 1.206 & 0.811 \\
    \midrule
    AvatarForcing* & Yes & 0.080 & 0.145 & 27.78 & \cellcolor{significantgray}5.891
      & 4.887 & \cellcolor{significantgray}3.917 & 1.337 & 1.800 \\
    DyStream & Yes & 0.101 & 0.147 & 33.32 & 7.106
      & 4.731 & 3.840 & 1.300 & \cellcolor{significantgray}1.673 \\
    AVTR-1 (Ours) & Yes & \cellcolor{significantgray}0.063 & \cellcolor{significantgray}0.136 & \cellcolor{significantgray}19.85 & 5.924
      & \cellcolor{significantgray}5.096 & 3.478 & \cellcolor{significantgray}1.432 & 0.900 \\
    \midrule
    Ground truth &  & 0.000 & 0.000 & 0.00 & 0.000
      & 5.251 & 4.116 & 1.552 & 1.700 \\
    \bottomrule
  \end{tabular}
\end{table}

\begin{table}[ht]
  \centering
  \small
  \setlength{\tabcolsep}{3pt}
  \caption{Conventional motion-based metrics computed from LivePortrait features on all frames of the SI-184 subset.
  For metrics without an arrow, values closer to ground truth are better.
  Expression PFD and variance are reported in units of $10^{-3}$ and $10^{-5}$, respectively.
  Their rotation components are reported in units of $10^{-2}$.
  Gray marks the best value among dyadic systems.}
  \label{tab:motion-metrics-liveportrait-full}
  \begin{tabular}{lcrrrrrrrr}
    \toprule
      & & \multicolumn{2}{c}{rPCC $\downarrow$} & \multicolumn{2}{c}{PFD $\downarrow$}
      & \multicolumn{2}{c}{SID} & \multicolumn{2}{c}{Var} \\
    \cmidrule(lr){3-4}\cmidrule(lr){5-6}\cmidrule(lr){7-8}\cmidrule(lr){9-10}
    Method & Paired audio & Exp & Rot & Exp & Rot & Exp & Rot & Exp & Rot \\
    \midrule
    SoulX Lite & No & 0.092 & 0.179 & 0.554 & 3.867
      & 5.021 & 3.133 & 3.400 & 0.854 \\
    SoulX Pro & No & 0.091 & 0.154 & 0.521 & 3.089
      & 5.070 & 3.309 & 3.500 & 1.044 \\
    Ditto & No & 0.123 & 0.177 & 0.836 & 4.300
      & 5.012 & 3.027 & 3.700 & 1.004 \\
    FLOAT & No & 0.129 & 0.144 & 0.686 & 2.764
      & 4.830 & 3.252 & 3.800 & 0.865 \\
    \midrule
    AvatarForcing* & Yes & 0.143 & \cellcolor{significantgray}0.152 & 1.043 & 4.653
      & 4.854 & \cellcolor{significantgray}3.830 & 4.700 & 2.947 \\
    DyStream & Yes & 0.168 & 0.172 & 1.155 & 4.349
      & 4.544 & 3.458 & \cellcolor{significantgray}3.900 & \cellcolor{significantgray}1.212 \\
    AVTR-1 (Ours) & Yes & \cellcolor{significantgray}0.110 & 0.165 & \cellcolor{significantgray}0.567 & \cellcolor{significantgray}3.824
      & \cellcolor{significantgray}5.043 & 3.191 & 3.200 & 0.922 \\
    \midrule
    Ground truth &  & 0.000 & 0.000 & 0.000 & 0.000
      & 5.219 & 3.862 & 4.000 & 2.027 \\
    \bottomrule
  \end{tabular}
\end{table}

\FloatBarrier

\section{R-DGG Temporal-Offset Analysis}
\label{app:temporal-offset-analysis}

\begin{figure}[H]
  \centering
  \includegraphics[width=0.92\linewidth]{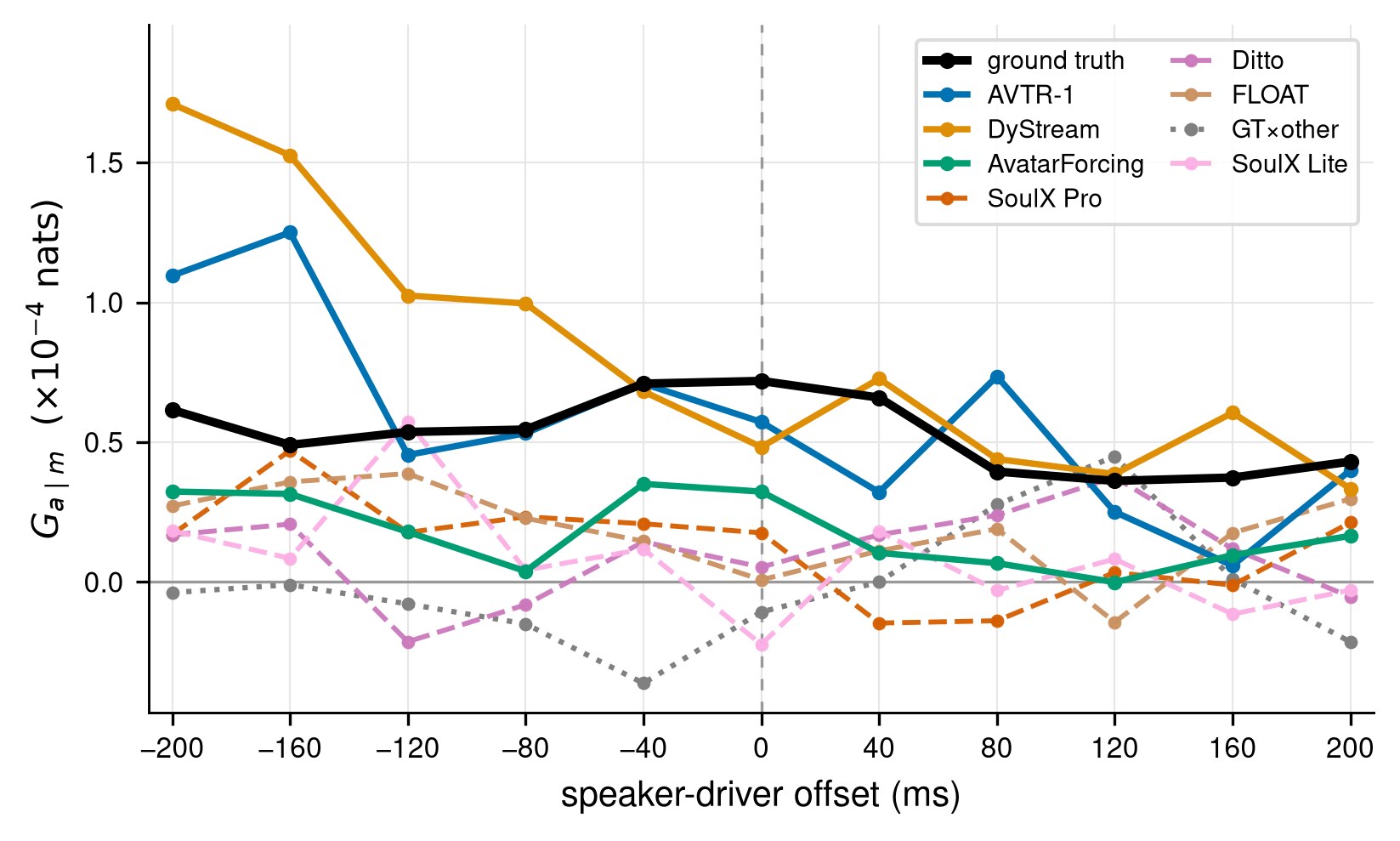}
  \caption{Sensitivity of R-DGG to a joint temporal offset of the speaker's motion and speech.
  The value at zero offset is reported in Table~\ref{tab:granger-results}.}
  \label{fig:granger-offsets}
\end{figure}

Figure~\ref{fig:granger-offsets} shows that the gain changes with temporal alignment for ground truth and the
three dyadic models, while the talking-head generators and GT~$\times$~other remain approximately flat.
This pattern is consistent with the distinction observed in Table~\ref{tab:granger-results}.
Because the individual offsets have no bootstrap intervals, we do not interpret the maxima as response latencies
or use them to rank the models.

\FloatBarrier